%% file: arxiv.tex
\documentclass{article}

\usepackage[preprint]{thesis} 
\usepackage{mathrsfs}
\usepackage{algorithmic}
\usepackage{algorithm}
\usepackage{amsmath}
\usepackage{amssymb}
\usepackage{booktabs}
\usepackage{graphicx}
\usepackage{subcaption}
\usepackage{enumitem}
\usepackage{siunitx}
\usepackage{gensymb}
\usepackage{svg}
\usepackage[capitalize]{cleveref}
\usepackage{placeins}
\usepackage{multirow}
\usepackage[table,dvipsnames,svgnames,x11names]{xcolor}
\usepackage{hyperref}

\title{Smooth Operator: A Real-Time Sampling-Based Algorithm for Kinematic Hand Retargeting}

\author{
  Robert Jomar Malate$^{1, 3, *}$ \quad
  Erik Bauer$^{2}$ \quad
  Norica Bacuieti$^{3}$ \quad
  Stefanos Charalambous$^{3}$ \\
  \textbf{Elvis Nava}$^{3}$ \quad 
  \textbf{Robert K. Katzschmann}$^{1, 3, *}$ \quad
  \textbf{Benedek Forrai}$^{3, *}$ \\
  \\
  $^1$ETH Zurich, Switzerland \quad
  $^2$Stanford University, United States \quad
  $^3$mimic robotics, Switzerland \\
  \\
  {\small * Corresponding authors: \href{mailto:robjmal@gmail.com}{\texttt{rmalate@student.ethz.ch}},} \\
  {\small \href{mailto:benedek.forrai@mimicrobotics.com}{\texttt{benedek.forrai@mimicrobotics.com}}, and \href{mailto:rkk@ethz.ch}{\texttt{rkk@ethz.ch}}} \\
}

\begin{document}
\newcommand{\numparticipants}{18}
\newcommand{\numrealworldtasks}{3}
\maketitle

\vspace{-3.0em}
\begin{center}
\url{https://mimicrobotics.github.io/smooth-operator/}
\end{center}


\begin{figure}[htbp]
\centering
\includegraphics[width=1.0\linewidth]{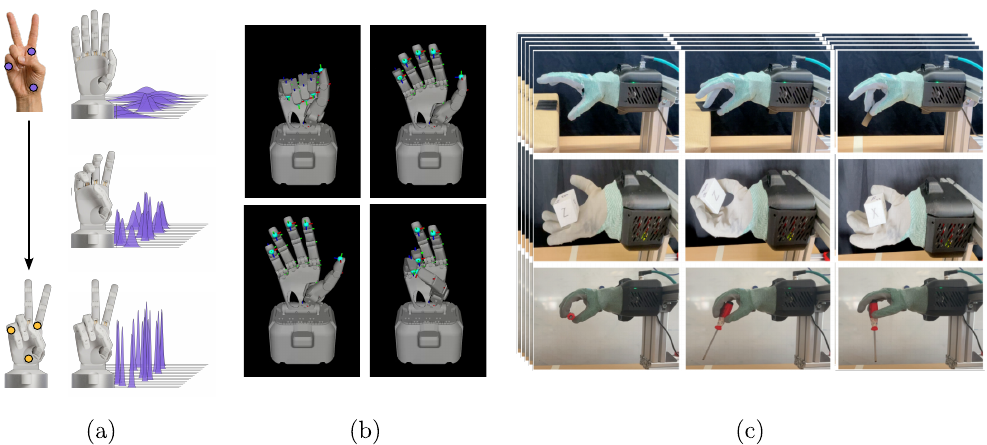}
\caption{\textbf{Overview.} Drawing from the rich traditions of sampling-based control in robotics, we introduce a sampling-based retargeter for tracking human hand pose commands with dexterous robotic hands (a). We rigorously evaluate its performance against the state of the art along several metrics, both in a simulated environment (b) that allows quick iteration, and in a real-world user study involving \numparticipants{} participants on \numrealworldtasks{} complex manipulation tasks (c). We find that our novel sampling based method leads to a better user experience, reduced jitter, and achieves state of the art performance in task success rates and operator workload scores.}
\label{fig:benchmark_hands_visualization}
\end{figure}

\begin{abstract}
    Advances in learning-based robotic manipulation, such as Vision-Language-Action (VLA) models and Video Action Models (VAMs), heavily rely on high-quality teleoperation data. Their capabilities are strictly upper-bounded by the quality of the underlying human demonstrations. Current gradient-based retargeting algorithms often converge to different local minima, resulting in jitter that affects data quality and teleoperation experience. To address this, we introduce the \textbf{Sampling-Based Retargeter (SBR)}, a novel gradient-free retargeting method drawn from the rich literature of sampling-based control and explicitly designed for low-jitter, real-time kinematic retargeting. We evaluate SBR both in simulation and through a rigorous real-world user study involving \numparticipants{} participants performing \numrealworldtasks{} complex manipulation tasks. Compared to gradient-based baselines, SBR achieved the highest overall task success rate (54.1\%) while significantly reducing operator cognitive fatigue, recording the lowest NASA-TLX workload score (36.4 out of 100). Ultimately, we establish SBR as a highly effective, intuitive retargeter for dexterous manipulation, providing the community with a rigorous benchmarking methodology to guide future retargeting research.
\end{abstract}


\input{thesis_smooth_operator/sections/01_introduction}

\input{thesis_smooth_operator/sections/02_related_works}
\input{thesis_smooth_operator/sections/03_sampling_based_retargeter}
\input{thesis_smooth_operator/sections/04_offline_analysis}
\input{thesis_smooth_operator/sections/05_real_world_experiments}

\input{thesis_smooth_operator/sections/06_discussion}

\input{thesis_smooth_operator/sections/07_limitations}

\input{thesis_smooth_operator/sections/08_conclusion}

\clearpage

\acknowledgments{
We thank mimic robotics for providing the hardware and simulation assets, and we extend our gratitude to the employees who participated in the experiments. Additionally, we thank the Soft Robotics Lab for providing feedback throughout the project. Erik Bauer is supported by a Stanford Graduate Fellowship (SAP Fellowship Fund), and thanks Shuran Song for feedback in early stages of the project.
}

\bibliography{arxiv}  

\clearpage
\appendix 

\begin{center}
    \LARGE \bf Supplementary Material
\end{center}
\vspace{2em}

\input{thesis_smooth_operator/appendix/01_sbr_hyperparams}
\FloatBarrier 

\input{thesis_smooth_operator/appendix/02_experiment_design}
\FloatBarrier

\input{thesis_smooth_operator/appendix/03_operator_adaptation_effects}
\FloatBarrier 

\input{thesis_smooth_operator/appendix/04_operator_variance}
\FloatBarrier 

\input{thesis_smooth_operator/appendix/05_gradient_v_sampling}
\FloatBarrier 

\input{thesis_smooth_operator/appendix/06_sim_to_real}

\end{document}

%% file: thesis_smooth_operator/sections/01_introduction.tex
\section{Introduction}
\label{sec:introduction}
Retargeting algorithms enable humans to bridge the human-robot embodiment gap, providing a method to control and teach robots to perform dexterous behaviors autonomously. Defined as the mapping of intent from a source to a target~\cite{Sivakumar2022_RoboticTelekinesis}, retargeting is a critical component of the robot learning data flywheel, providing a mechanism for extracting human manipulation data through real-time control~\cite{Qin2022_DexMV, Qin2024_AnyTeleop, Nava2025_mimicone} or offline processing~\cite{Mandi2025_DexMachina, Pan2026_SPIDER}. We call the former and latter \textit{online} and \textit{offline retargeting}, respectively. However, retargeting is challenging due to the complex and nonlinear mapping between the human and robot hand manifolds~\cite{Yin2025_GeoRT}. This complexity arises from the kinematic discrepancies between the embodiment gap. To address the embodiment gap, a wide variety of approaches have been developed to translate human hand motion into executable robot commands~\cite{Meattini2023_RetargetingLitReview}. 

These retargeting approaches commonly use gradient-based optimization, which relies on local derivative information to map human intent to robot joints. Gradient solvers, however, risk converging into undesirable local minima, thereby generating aggressive, high-frequency command spikes (jitter). In real-world teleoperation, this erratic behavior degrades grasp stability and makes aggressive low-pass filtering necessary, reducing the teleoperation bandwidth of the whole system. 

Furthermore, there is no clear way to rigorously evaluate the performance of retargeting algorithms. Previous works often used very small sample sizes of operators~\cite{Handa2019_DexPilot, Sivakumar2022_RoboticTelekinesis, Naughton2024_ResPilot, Yin2025_GeoRT}, compromising statistical rigor for A/B comparisons. Additionally, they do not properly account for operator adaptation with the tasks and retargeters, which violates typically assumed statistical independence for sequential testing of multiple retargeters. Moreover, the evaluation setup of some works~\cite{Sivakumar2022_RoboticTelekinesis, Qin2024_AnyTeleop, Ding2024_BunnyVisionPro, Xin2025_AnalyzingKeyObjectives} robot hand and arm control. Arm control is a confounding factor for evaluating retargeting methods for hands, as it allows users to compensate for poor hand retargeting by adapting their strategy for arm movement. This confounds the evaluation by making it difficult to isolate the actual contribution of the retargeting algorithm to capturing the operator's intent. 

The main contribution of this work is a novel gradient-free, sampling-based retargeter (SBR) for real-time hand teleoperation. SBR evaluates a broad distribution of candidate joint configurations of the robot hand and updates its control distribution through a cost-weighted softmax over elite samples, probabilistically bounding trajectory variance. As a result of bounded trajectory variance, the optimizer acts as an inherent kinematic low-pass filter, producing smooth, low-jitter commands without post-hoc filtering. To our knowledge, this is the first application of a sampling-based method to real-time human-hand retargeting. We validate SBR through a rigorous user study of \numparticipants{} operators across three dexterous manipulation tasks, applying Balanced Latin squares~\cite{Bradely1958_BalancedLatinSquares} with mixed-effects modeling~\cite{Seabold2010_Statsmodels, Gelman2006_DataAnalysisMEMs, Baayen2008_MixedEffectsModeling} to account for operator learning, and isolating retargeter fidelity via a static-fixture hand mount. Our results demonstrate that smoother retargeting directly translates to higher task success rates and lower operator cognitive load, establishing SBR as a practical, scalable foundation for data collection of dexterous behaviors.

%% file: thesis_smooth_operator/sections/02_related_works.tex
\section{Related Work}
\label{sec:related_work}

\paragraph{Online Retargeting}
\label{subsec:related_works_online_retargeting}
Various approaches to online retargeting and their respective benchmarking methodologies have been proposed in the literature ~\cite{Meattini2023_RetargetingLitReview}. Early approaches to online retargeting relied on direct joint angle mapping~\cite{Li2022_SurveyOnMapping, Meattini2023_RetargetingLitReview}, but these methods often fail to preserve spatial task geometry due to the kinematic discrepancies between human and robot hands. To address this, recent systems have adopted Human Keyvector Matching (HKVM)~\cite{Handa2019_DexPilot, Qin2024_AnyTeleop}, utilizing heuristics~\cite{Handa2019_DexPilot}, Gaussian Processes~\cite{Naughton2024_ResPilot}, and analyzing key objectives with additional loss terms~\cite{Xin2025_AnalyzingKeyObjectives} to improve grasping and finger-gaiting performance. Alternatively, learning-based approaches have modeled these nonlinear mappings via supervised autoencoders~\cite{Chong2021_LearningBasedHarmonicMapping} or unsupervised geometric objectives~\cite{Yin2025_GeoRT}.


\paragraph{Sampling-Based Control}
\label{subsec:related_works_sampling_based_control}
Sampling-based control provides a robust alternative to gradient-based solvers in non-convex or non-differentiable domains~\cite{Lavalle1998_RRT, Williams2016_MPPI}, finding applications in robot navigation and locomotion~\cite{Howell2022_PredictiveSamplingMujoco, Keshavarz2025_Controlleggedrobotsusing}. Recently, predictive sampling has been extended to dexterous manipulation to navigate high-dimensional, contact-rich state spaces in both simulation~\cite{Howell2022_PredictiveSamplingMujoco} and dynamic physical hardware deployments~\cite{Hess2025_SamplingBasedMPC, Zhang2026_Sumo}. However, its application to motion retargeting remains largely unexplored. While DynaRetarget~\cite{Dhedin2026_DynaRetarget} successfully utilizes sampling optimization for humanoid retargeting, their approach is strictly designed for offline processing and locomotion, where computational latency is not a primary constraint. No prior method achieves real-time sampling-based retargeting for human hand control. 

%% file: thesis_smooth_operator/sections/03_sampling_based_retargeter.tex
\section{Sampling-Based Retargeter}
\label{sec:sampling_based_retargeter}

\begin{figure}[htbp]
\label{fig:sbr_system_overview}
\centering
\includegraphics[width=1.0\linewidth]{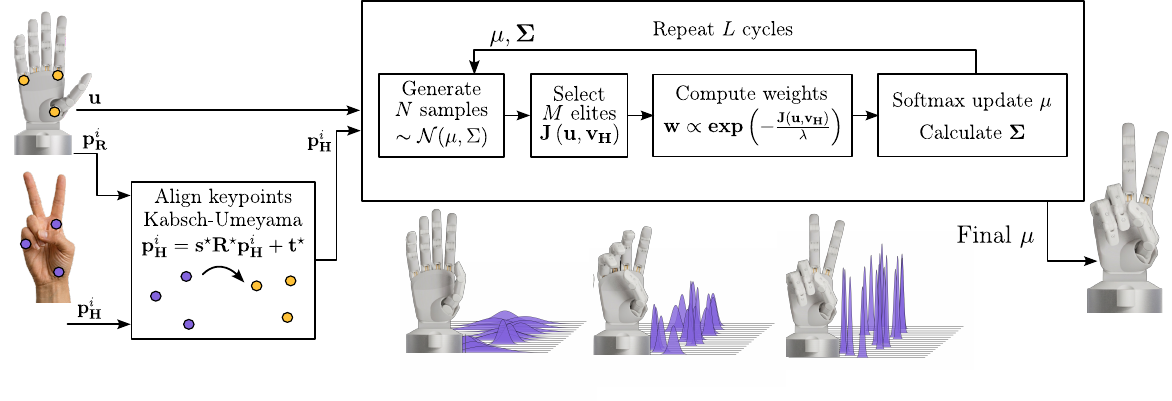}
\caption{\textbf{System overview.} After aligning human ($\mathbf{p}_H^i$) and robot keypoints using the Kabsch-Umeyama algorithm~\cite{Kabsch1976, Kabsch1978, Umeyama1991}, SBR optimizes joint commands iteratively. Over $L$ cycles, $N$ candidates are sampled in each iteration, and $M$ elite samples are weighted via a softmax function to update the control distribution ($\mu, \Sigma$) subject to the cost function $J$~(\cref{eq:sbr_loss_function}).}
\end{figure}
To overcome the limitations of gradient-based solvers, we frame retargeting as a gradient-free optimization problem via our SBR. It utilizes Model Predictive Optimized Path Integral (MPOPI)~\cite{Asmar2023_MPOPI} framework for control stability while using the Improved Cross-Entropy Method (iCEM)~\cite{Pinneri2020_iCEM} as the Adaptive Importance Sampling (AIS) method. Rather than relying on noisy local gradients, SBR evaluates a wide distribution of candidate joint configurations. By isolating high-performing ``elite`` samples and applying a cost-weighted softmax function to update the control distribution, SBR probabilistically bounds trajectory variance. This allows the optimizer to act as an inherent kinematic low-pass filter, yielding the smooth, low-jitter motion required for physical teleoperation. To achieve sufficient sample throughput for real-time teleoperation, we implement our framework in JAX~\cite{jax2018github} and MJX forward kinematics~\cite{Zakka2022_MujocoMenagerie}, which enable efficient parallelization of forward kinematics and cost evaluations.

\paragraph{Formulation}
\label{sec:sbr_formulation}
We define the state of the robot hand $q_{t} \in \mathbb{R}^{d}$ at timestep $t$ as a vector of its $d$ joint angles. The next state $q_{t+1}$ is equal to the retargeter output $u_{t}$. The objective is to find a $u_{t}$ that minimizes a composite cost function $J$, which is a modified version of DexPilot's loss function~\cite{Handa2019_DexPilot} with a temporal regularization term to promote smoothness \cite{Xin2025_AnalyzingKeyObjectives}: 

\begin{equation}
\label{eq:sbr_loss_function}
    J = \sum_{k=1}^{K} \lambda_{k}|| v_{R}^{k} - v_{H}^{k}||^2 + \lambda_{reg}|| q_{R} - q_{R,neutral}||^2 + \lambda_{vel}|| q_{R}^{} - q_{R,prev} ||^2
\end{equation}

where $K$ is the number of keyvectors, $q_{R}$ is the joint angles of the robot hand and $q_{H}$ is the pose of the human hand. $v_{H}^{k}$ and $v_{R}^{k}$ represent the human and robot keyvectors, respectively. $q_{R, neutral}$ is the neutral hand pose (defined as zero angles) and $q_{R,prev}$ is the state of the robot hand in the previous timestep. $\lambda_k$ is the weight for the $k$-th keyvector term, while $\lambda_{reg}$ and $\lambda_{vel}$ represent the weights of the pose regularization and velocity regularization terms, respectively. The full algorithm is detailed in~\Cref{algo:sampling_based_retargeter}.

\begin{algorithm}
\caption{Sampling-Based Retargeter}
\begin{algorithmic}[1]
\label{algo:sampling_based_retargeter}
\REQUIRE Initial state $q$, Initial covariance $\sigma$, number of samples $N$, number of elite samples $M$, temperature parameter $\lambda$, update cycle $L$, numerical stability term $\epsilon$, joint limits $[q_{min}, q_{max}]$
\ENSURE Optimal control input $u$

\STATE Initialize mean $\mu$ and covariance $\Sigma$: $\mu \leftarrow q$, $\Sigma \leftarrow \operatorname{diag(\sigma^{2})} + \epsilon I$
\STATE Initialize update mean and covariance variables: $\mu' \leftarrow \mu$, $\Sigma' \leftarrow \Sigma$

\FOR{each $l = 1, \dots, L$}
    \STATE Sample N control candidates: $u_{n} \sim \mathcal{N}(\mu', \Sigma')$
    \STATE Compute cost for each candidate: $J_{n}(u_{n})$

    \IF{$l < L$}
        \STATE Identify $M$ elite samples $\mathcal{U}_E = \{u_m\}_{m=1}^M$ with lowest costs $\{J_m\}_{m=1}^M$
        \STATE Find minimum elite loss: $J_{min} = \min(\{J_m\})$
        \STATE Compute Softmax weights: $w_{m} = \frac{\exp\left(-\frac{J_{m} - J_{min}}{\lambda}\right)}{\sum_{i=1}^{M}\exp\left(-\frac{J_{i} - J_{min}}{\lambda}\right)}$
        \STATE Update mean: $\mu' \leftarrow \sum_{m=1}^{M} w_{m} u_{m}$
        \STATE Update covariance using new mean: $\Sigma' \leftarrow \sum_{m=1}^{M} w_{m} (u_{m} - \mu')(u_{m} - \mu')^\top + \epsilon I$
    \ENDIF
\ENDFOR

\STATE Apply joint limit clipping: $u \leftarrow \operatorname{clip}(\mu', q_{min}, q_{max})$
\STATE Update state: $q \leftarrow u$
\STATE Apply control output: $u$

\end{algorithmic}
\end{algorithm}

%% file: thesis_smooth_operator/sections/04_offline_analysis.tex
\section{Offline Analysis}
\label{sec:benchmark}

\begin{figure}[htbp]
\centering
\includegraphics[width=1.0\linewidth]{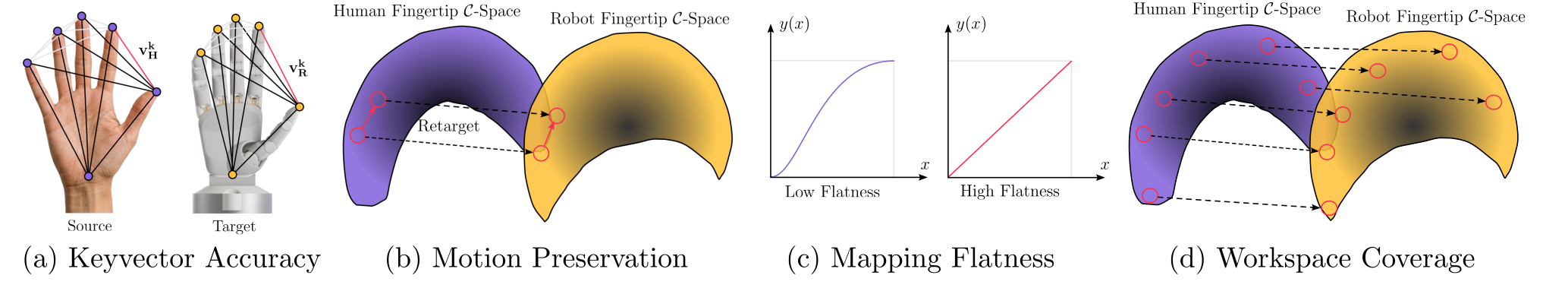}
\caption{\textbf{Heuristic metrics for human-to-robot hand motion retargeting.} Individual panels show geometric accuracy (3a), trajectory alignment (3b), smooth mapping (3c), and total workspace coverage optimization (3d). Figures 3b, 3c, and 3d are reformatted from~\cite{Yin2025_GeoRT}.}
\label{fig:retargeting_metrics}
\end{figure}

\paragraph{Retargeting Heuristics}
\label{subsec:offline_analysis_retargeting_heursitics}
To help us tune the loss function, we measure the performance of the following metrics that are used in the retargeting literature. We hypothesize that SBR's sampling-based approach will yield superior consistency metrics while remaining competitive in accuracy. Refer to \cref{sec:sim_to_real_metrics} for a more in-depth explanation of these metrics:

\begin{itemize}[leftmargin=*]
    \item \textbf{Accuracy:} Keyvectors capture the intended spatial relationship between embodiments~\cite{Handa2019_DexPilot, Qin2024_AnyTeleop} (\cref{fig:retargeting_metrics}a). We assess the directional alignment of corresponding human ($v_{H}^{k}$) and robot keyvectors ($v_{R}^{k}$) via \underline{Cosine Similarity}. \underline{Scale Ratio} measures the spatial magnitude preservation.

    \item \textbf{Consistency:} This metric quantifies the kinematic stability, critical for operator intuitiveness ~\cite{Yin2025_GeoRT}. \underline{Motion Preservation} assesses the directional tracking of human and robot landmark velocities (~\cref{fig:retargeting_metrics}b). \underline{Flatness} penalizes high-frequency control jitter(~\cref{fig:retargeting_metrics}c).
    
    \item \textbf{Efficiency:} We evaluate the retargeter's ability to leverage its joint range while maintaining real-time frequencies. \underline{Workspace Utilization} measures the volumetric overlap between the reachable workspace ~\cite{Yin2025_GeoRT, Naughton2024_ResPilot} (\cref{fig:retargeting_metrics}d). 
\end{itemize}


We execute these benchmarks within a MuJoCo simulation environment, using prerecorded human hand data from MANUS Metagloves Pro~\cite{ManusMetaglovesPro} as the driving input. To ensure a fair comparison on a standardized morphology, we benchmark SBR against three open-source, state-of-the-art retargeters operating on the \textit{mimic P0.50} hand~\cite{Nava2025_mimicone}: 

\begin{enumerate}[label=\alph*)]
\item \textbf{DexPilot}~\cite{Handa2019_DexPilot}: A heuristic-based HKVM approach that popularized keyvector matching for teleoperation. It serves as the primary standard in our evaluation~\cite{Sivakumar2022_RoboticTelekinesis, Qin2024_AnyTeleop, Naughton2024_ResPilot, Yin2025_GeoRT}.
\item \textbf{Hybrid}~\cite{Nava2025_mimicone}: A HKVM approach that uses joint angle information to warmstart the optimization procedure. 
\item \textbf{Geometric Retargeter (GeoRT)}~\cite{Yin2025_GeoRT}: An unsupervised learning-based approach that utilizes geometric objective functions.
\item \textbf{Sampling-Based Retargeter (Ours)}: Our proposed method, detailed in~\cref{sec:sampling_based_retargeter}.
\end{enumerate}

\begin{table}[htbp]
    \centering
    \small
    \begin{tabular}{@{}l l c c c c@{}}
    \toprule
    \textbf{Category} & \textbf{Metric} & \textbf{DexPilot}~\cite{Handa2019_DexPilot} & \textbf{GeoRT}~\cite{Yin2025_GeoRT} & \textbf{Hybrid}~\cite{Nava2025_mimicone} & \textbf{SBR (Ours)} \\
    \midrule
    \multirow{2}{*}{\textbf{Accuracy}}
        & Cos. Similarity ($\rightarrow 1$) & 0.9932 & 0.9799 & 0.9879 & \textbf{0.9969} \\
        & Scale Ratio ($\rightarrow 1$)     & 0.885  & \textbf{1.050} & 0.891 & 0.877 \\
    \midrule
    \multirow{2}{*}{\textbf{Consistency}}
        & Motion Preserv. ($\rightarrow 1$) & 0.727 & 0.695 & 0.675 & \textbf{0.776} \\
        & Flatness ($\downarrow$)           & 7.71e-06 & 3.01e-06 & 6.17e-06 & \textbf{7.17e-07} \\
    \midrule
    \multirow{2}{*}{\textbf{Efficiency}}
        & Workspace Util. ($\uparrow$) & 26.1\% & \textbf{35.2\%} & 22.5\% & 21.0\% \\
        & Latency [ms] ($\downarrow$)  & 4.905 & \textbf{1.410} & 9.892 & 3.664 \\
    \bottomrule
    \end{tabular}
    \vspace{0.1cm}
    \caption{\textbf{Simulation benchmark results for online retargeting algorithms.} Arrows indicate the direction of optimality and best performing values are bolded. SBR achieves the highest Consistency metrics and Cosine Similarity. Data from each category is averaged over $N=14$ operators. Scheduling constraints limited the analysis to 14 of the 18 operators.}
    \label{table:benchmark_results}
\end{table}

As demonstrated in~\cref{table:benchmark_results}, SBR establishes a new state-of-the-art in kinematic consistency. By achieving the highest Motion Preservation ($0.776$) and an order-of-magnitude reduction in Flatness ($7.17\text{e-}07$), SBR validates its design as a temporal low-pass filter that effectively eliminates high-frequency control jitter. Furthermore, SBR outperforms the other retargeters in Cosine Similarity ($0.9969$), demonstrating that it maintains superior geometric intent without relying on hand-crafted heuristics. This robust kinematic consistency motivates our evaluation of SBR's real-world performance in the subsequent teleoperation study (\cref{sec:real_world_experiments}).

%% file: thesis_smooth_operator/sections/05_real_world_experiments.tex
\section{Real World Experiments And Results}
\label{sec:real_world_experiments}
To evaluate real-world retargeting performance, we conduct a user study utilizing the \textit{mimic P0.50} hand~\cite{MimicRobotics2026_mimicP050Hand}, teleoperated via MANUS Metagloves Pro~\cite{ManusMetaglovesPro}. We aim to validate the following hypothesis: the improved consistency of the SBR leads to improved teleoperation success rates and better user experience. Refer to \cref{sec:experiment_design} for a more detailed explanation of the experimental design.

\paragraph{Setup}
\label{sec:rw_exp_setup}
Following the evaluation protocol established by Naughton et al.~\cite{Naughton2024_ResPilot}, the robotic hand is attached to a static fixture. This isolates the retargeter's mapping fidelity by forcing operators to rely entirely on finger articulation, preventing them from using compensatory robotic arm movements to complete the objective. Due to hardware limitations, the control rate for all retargeters is fixed at 30\si{\hertz}. Using this setup, we select three distinct finger-gaiting tasks directly adapted from Naughton et al.~\cite{Naughton2024_ResPilot}, each representing a different axis of dexterous manipulation:
\begin{enumerate}[label=\alph*)]
    \item \textit{Card Pickup (Task A)}: The operator retrieves a flat card from the table by sliding it over the edge and pinching it.
    \item \textit{Vertical Cube Rotation (Task B)}: The operator rotates a cube 90 degrees about the vertical axis. This requires complex finger-gaiting and strictly tests how well the algorithm translates high-dexterity coordination from the human to the robot.
    \item \textit{Screwdriver Pivot (Task C)}: The operator grasps a screwdriver with 4 of the fingers. Afterwards, they slowly loosen their grip and adjust the screwdriver until it points downwards. This evaluates precision control and basic contact stability.
\end{enumerate}

\begin{figure}[htbp]
\centering
\includegraphics[width=0.8\linewidth]{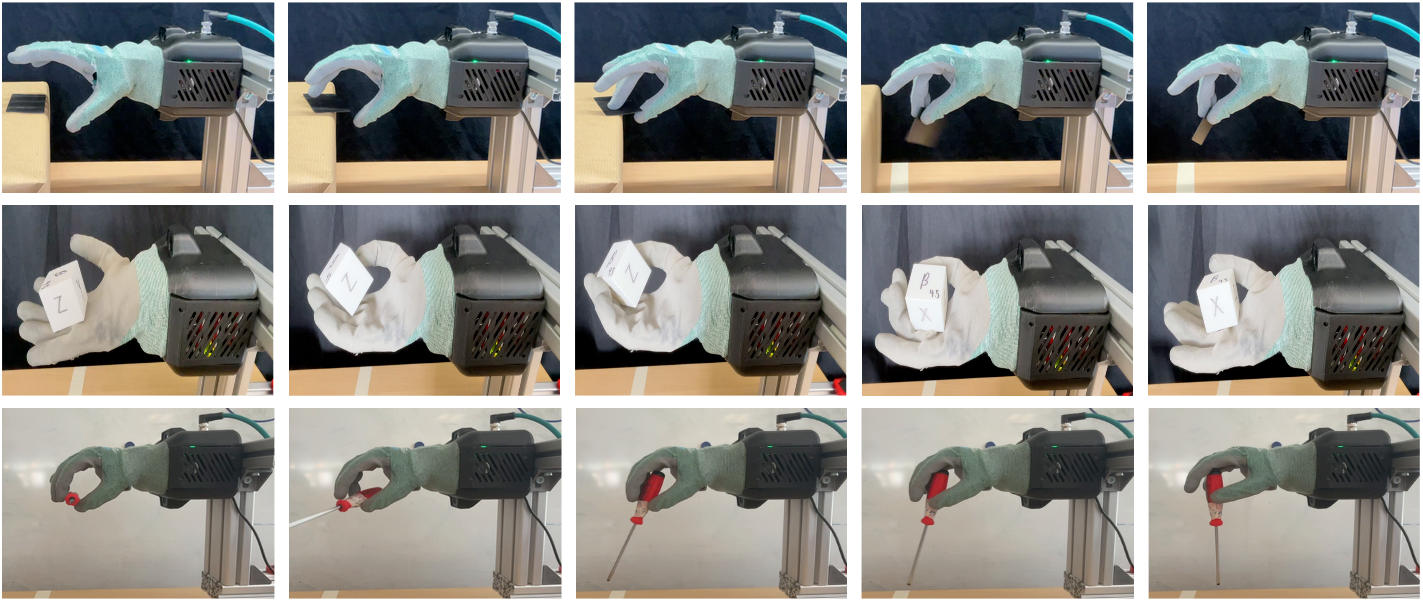}
\caption{\textbf{Real world tasks.} Highly dexterous tasks used in real world experiments to benchmark the tasks: \textit{Card Pickup (Task A, top)}, \textit{Vertical Cube Rotation (Task B, middle)}, and \textit{Screwdriver Pivot (Task C, bottom)}.}
\label{fig:rw_tasks}
\end{figure}

Our study is comprised of a cohort of \numparticipants{} participants (three of whom are paper authors). The cohort possesses varying levels of teleoperation experience, ranging from experts to a majority of complete novices. This distribution ensures that our evaluation captures the generalizability and intuitiveness of the retargeters for standard operators during large-scale data collection. Participants test four algorithms, i.e., DexPilot~\cite{Handa2019_DexPilot}, GeoRT~\cite{Yin2025_GeoRT}, Hybrid~\cite{Nava2025_mimicone}, and SBR (Ours), in isolated task blocks. Scheduling constraints yielded varying sample sizes per task (Task A: $N=12$, B: $N=8$, C: $N=6$). Crucially, to evaluate true out-of-the-box performance, the algorithms were \textit{not} manually tuned for individual operators; participants only underwent a standard baseline calibration for the MANUS gloves prior to testing. To mitigate order and learning effects, the algorithm sequence follows a Balanced Latin Square design~\cite{Bradely1958_BalancedLatinSquares}. For each assigned retargeter, the operator received a strict two-minute period where they can move the fingers freely \textit{without practicing the specific task}. After this time period, they begin performing consecutive recorded trials (10 trials for Tasks A and B; 20 trials for Task C).

Task performance is evaluated by measuring the completion time and a granular success score. 1.0 point is given for perfect task completion, while 0.5 points is for marginal success, representing the cases where the task was completed but the operator reached an undesirable intermediate state or required a reset. 0.0 points is given for task failure, defined as the object dropping out of the workspace. To quantify usability and cognitive burden, the NASA Task Load Index (TLX)~\cite{Hart1988_NASATLX} (scored 0--100) is administered at three points during each algorithm block: after the initial, midpoint, and final trials of each block. The subscale scores are averaged across all operators per task-retargeter pair. 

\paragraph{Results}
\label{sec:rw_exp_results_task_performance}
Teleoperation is an inherently adaptive process; as operators accumulate trial experience, they naturally exhibit both learning effects and progressive fatigue. Consequently, the statistical assumption of independence across consecutive trials does not hold. To rigorously evaluate retargeter performance while accounting for this intra-subject variance, we employ Mixed Effects Models (MEMs) via the \textit{statsmodels} library~\cite{Seabold2010_Statsmodels, Gelman2006_DataAnalysisMEMs, Baayen2008_MixedEffectsModeling}. For computing the estimated marginal means (EMM), we set DexPilot as the reference baseline; every reported $p_{\text{Dunnett}}$ tests the corresponding retargeter against this baseline, with Dunnett correction applied across the three resulting comparisons.

\begin{figure}[htbp]
    \centering
    \begin{subfigure}[b]{0.48\linewidth}
        \centering
        \centering
        \includegraphics[trim=0cm 0cm 0cm 0.5cm, clip, width=1.0\linewidth]{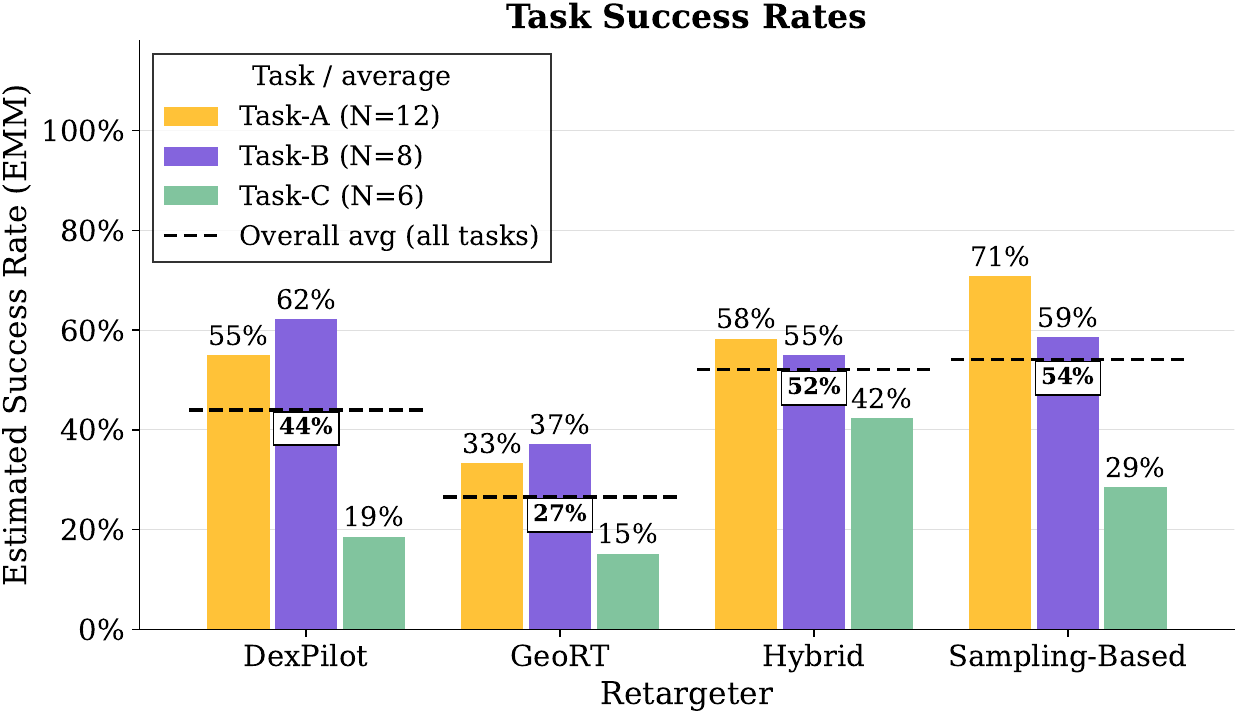}
        \caption{Task Success Rate}
        \label{fig:success_rate}
    \end{subfigure}
    \hfill 
    \begin{subfigure}[b]{0.48\linewidth}
        \centering
        \includegraphics[trim=0cm 0cm 0cm 0.6cm, clip, width=1.00\linewidth]{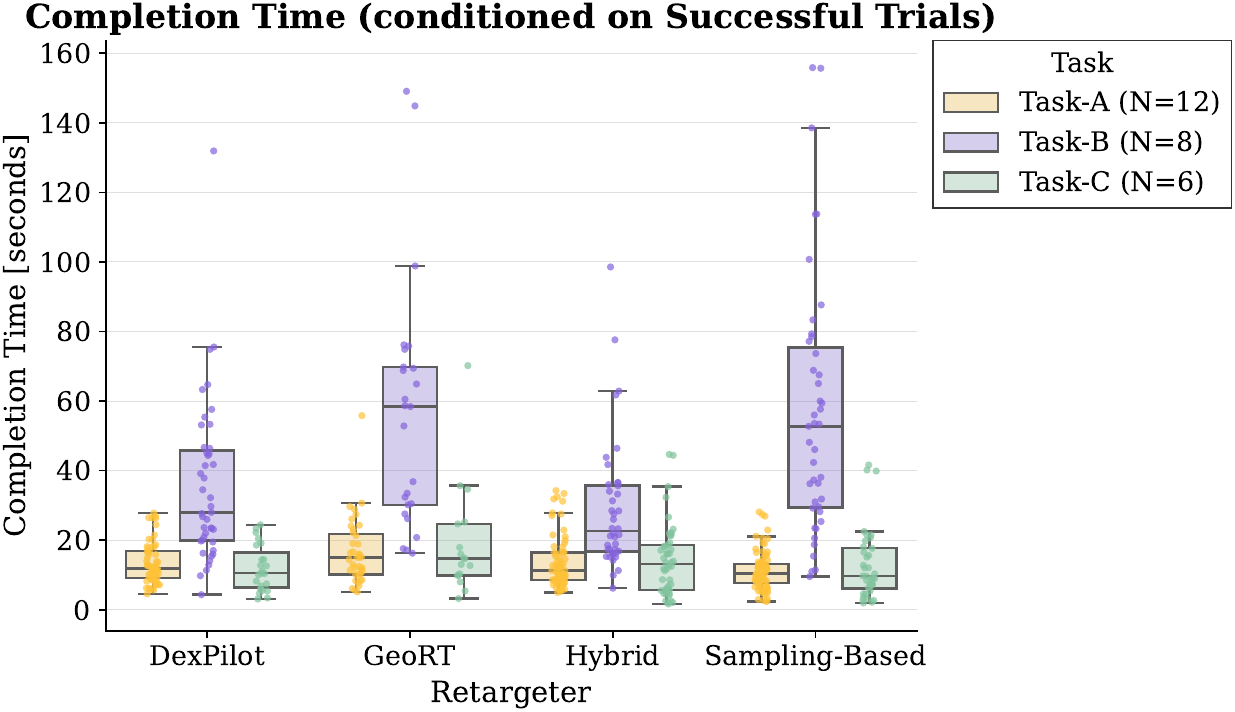}
        \caption{Task Completion Time}
        \label{fig:completion_time}
    \end{subfigure}
    
    \vspace{0.2cm} 
    \caption{\textbf{Empirical evaluation of retargeter performance.} (a) Compares the overall success rates across algorithms for Tasks A, B, and C. (b) Illustrates the variance and consistency in task completion times for successful trials.}
    \label{fig:empirical_performance}
\end{figure}
\begin{figure}[htbp]
\centering
\includegraphics[width=1.0\linewidth]{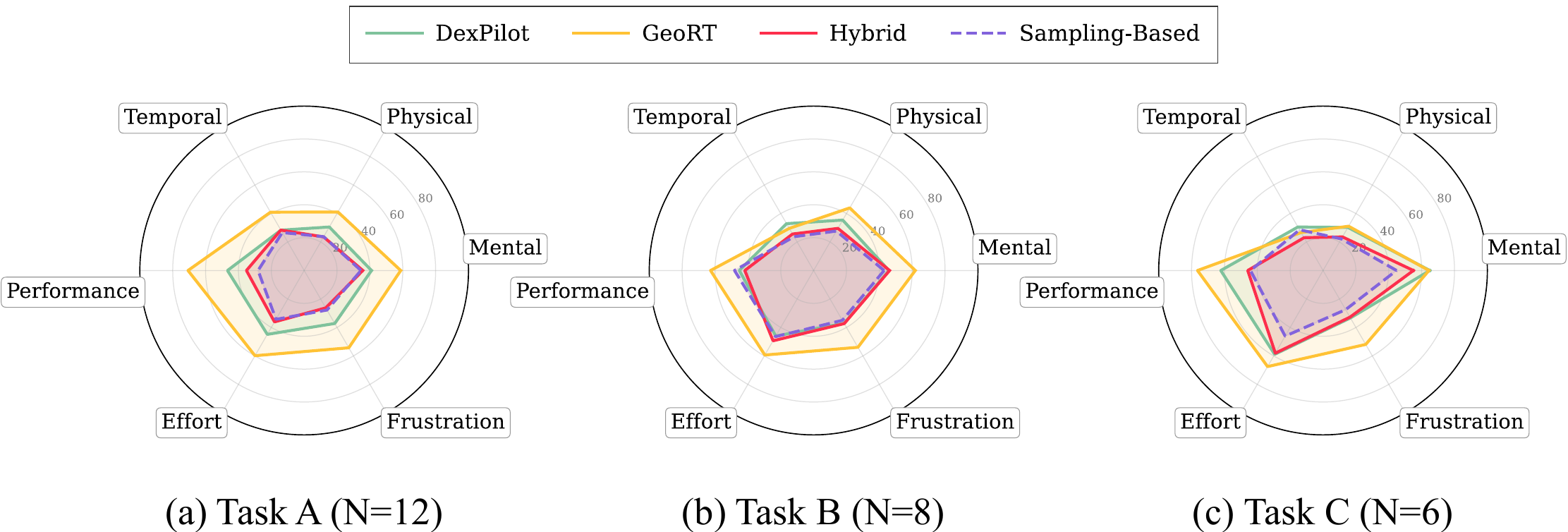}
\caption{\textbf{NASA TLX results for each task}.
Results from NASA-TLX survey from operators on the different retargeters for each task. A smaller area indicates that the retargeter had a reduced workload.}
\label{fig:rw_exp_results_nasa_tlx}
\end{figure}

\begin{itemize}[leftmargin=*]
    \item \textbf{Task Reliability (Success Rate):} SBR achieves the highest overall estimated marginal mean success rate of \textbf{54.1\%} ($p_{Dunnett}=0.038$). This exceeds the EMM of DexPilot (44.0\%), GeoRT ($26.6\%, p_{Dunnett}<0.001$), and Hybrid ($52.1\%, p_{Dunnett}=0.331$) (see~\cref{fig:empirical_performance}). This advantage was most pronounced in Task A, where SBR yielded a \textbf{70.9\%} success rate ($p_{Dunnett}=0.013$) compared to DexPilot (55.0\%), GeoRT (33.3\%, $p_{Dunnett}=0.001$), and Hybrid (58.3\%, $p_{Dunnett}=0.895$). 
    
    \item \textbf{Cognitive Usability (NASA-TLX):} SBR recorded the lowest overall cognitive workload score of \textbf{36.4} ($p_{Dunnett}<0.001$), showing a reduction compared to DexPilot (44.0), GeoRT (\textbf{56.4}, $p_{Dunnett}<0.001$) and Hybrid ($38.4, p_{Dunnett}=0.010$) (see~\cref{fig:rw_exp_results_nasa_tlx}).
    
    \item \textbf{Completion Time:} SBR had a slightly higher overall completion time of \textbf{29.7\si{\second}} ($p_{Dunnett}=0.013$) compared to DexPilot (23.3\si{\second}) and Hybrid (21.6\si{\second}, $p_{Dunnett}=0.805$). However, it is still faster than GeoRT (33.3\si{\second}, $p_{Dunnett}<0.001$).
\end{itemize}

%% file: thesis_smooth_operator/sections/06_discussion.tex
\section{Discussion}
\label{sec:discussion}
The real-world results strongly validate our core hypothesis: the improved smoothness of the sampling-based retargeter leads to competitive task success rates and reduced operator workload. State-of-the-art baselines using gradient-based optimization suffer from high-frequency jitters that our proposed method minimizes. Particularly for foundational grasp and finger-gaiting tasks, such as the card pickup (Task~A), reduced jitter plays a crucial role as evidenced by SBR's significantly improved task success rate ($+13\%$ over the closest baseline) and reduced workload.

\paragraph{Quantitative Results:} 
\label{subsec:quantitative_results}
We perform pairwise comparisons between all four retargeters. Reported significances ($p_{\text{Holm}}$) are Holm-corrected across the resulting six pairs within each scope. SBR achieves the highest pooled success rate of any retargeter ($54.1\%$), significantly above DexPilot ($44.0\%$, $p_{\text{Holm}}\!=\!0.042$) and GeoRT ($26.6\%$, $p_{\text{Holm}}\!<\!0.001$). It also achieves the lowest pooled workload ($36.4$ NASA-TLX points, $-7.6$ vs.\ DexPilot and $-20.1$ vs.\ GeoRT, both $p_{\text{Holm}}\!<\!0.001$). The differentiation is sharpest on Task~A, where SBR succeeds $70.9\%$ of the time vs.\ $55.0\%$ for DexPilot ($p_{\text{Holm}}\!=\!0.014$) and $33.3\%$ for GeoRT ($p_{\text{Holm}}\!<\!0.001$), while posting the fastest completion time ($12.3$~s) and lowest workload ($29.2$ pts) of any method on that task. SBR shows no significant per-task success degradation on Tasks~B and C, and significantly outperforms GeoRT on Task~B ($58.6\%$ vs.\ $37.2\%$, $p_{\text{Holm}}\!=\!0.013$). Against the Hybrid baseline, SBR matches both success and workload, with a consistent numerical edge in SBR's favor: lower workload in every scope (between $-1.0$ and $-4.6$ NASA-TLX points, all $p_{\text{Holm}}\!\geq\!0.30$) and higher success rate in three of four scopes ($+2.0$ percentage points pooled, $p_{\text{Holm}}\!=\!0.79$; $+12.6$ on Task~A, $p_{\text{Holm}}\!=\!0.16$; $+3.7$ on Task~B, $p_{\text{Holm}}\!=\!1.00$).

\paragraph{Qualitative Findings:} 
\label{subsec:qualitative_results}
Teleoperation runs confirm that SBR substantially reduces the erratic jittering inherited by all gradient-based baselines while maintaining precise tracking. Hybrid is particularly affected: its jitter is clearly visible during operation and was consistently flagged by test operators. By contrast, all 18 study participants described SBR as smooth, responsive, and predictable. This perceived stability, rather than any difference in raw task success, is the key factor separating SBR from Hybrid in operator preference.

%% file: thesis_smooth_operator/sections/07_limitations.tex
\section{Limitations}
\label{sec:limitations}
Despite its advantages, our evaluation framework presents several limitations. While our physical user study significantly advances the empirical rigor of dexterous retargeting evaluation compared to standard qualitative baselines, we still encountered constraints regarding statistical power. Teleoperation performance is inherently noisy; the combination of our specific sample size and the high variance in operators' baseline teleoperation experience introduces substantial inter-subject variability. Consequently, while broad performance trends are highly significant, detecting subtle interaction effects between operator skill and retargeter performance will require larger, systematically stratified cohorts in future work.

Additionally, a trade-off emerged regarding operational speed: successful SBR trials had a higher overall completion time (\SI{29.7}{\second}) than DexPilot (\SI{23.3}{\second}, $p_{\text{Holm}} = 0.014$) and Hybrid (\SI{21.6}{\second}, $p_{\text{Holm}} = 0.001$). This latency was localized to the complex rotation task (Task B), where SBR averaged \SI{61.9}{\second} against DexPilot's \SI{39.1}{\second} ($p_{\text{Holm}} = 0.001$). Future work will focus on closing this speed gap during dynamic sequences.

%% file: thesis_smooth_operator/sections/08_conclusion.tex
\section{Conclusion}
\label{sec:conclusion}

We introduced the Sampling-Based Retargeter (SBR), a novel gradient-free framework for low-jitter, real-time kinematic hand teleoperation. Rather than committing to a single local gradient update, SBR selects joint configurations from a cost-weighted distribution of candidates, making high-frequency command spikes far less likely than in gradient-based methods. The resulting trajectory smoothness maintains consistent contact during precision tasks, which is most evident in contact-rich manipulation such as card pickup (+13\% over the closest baseline). Across all tasks, SBR achieved a 54.1\% overall success rate, outperforming DexPilot (44.0\%) and GeoRT (26.6\%) and matching Hybrid ($52.1\%$), while reducing NASA-TLX workload to 36.4 points compared to DexPilot's (44.0), GeoRT (56.4), and Hybrid (38.4). Operator feedback consistently described SBR as exceptionally smooth, responsive, and intuitive across both novice and expert users. By making multi-finger coordination more predictable at the algorithmic level, SBR provides a principled and practical foundation for data collection of high-quality dexterous demonstrations.

%% file: thesis_smooth_operator/appendix/01_sbr_hyperparams.tex
\section{Sampling-Based Retargeter (SBR) Hyperparameters}
\label{sec:sampling_hyperparams}

\begin{figure}[htbp]
    \centering
    \includegraphics[trim=0cm 0cm 0cm 1.3cm, clip, width=0.95\linewidth]{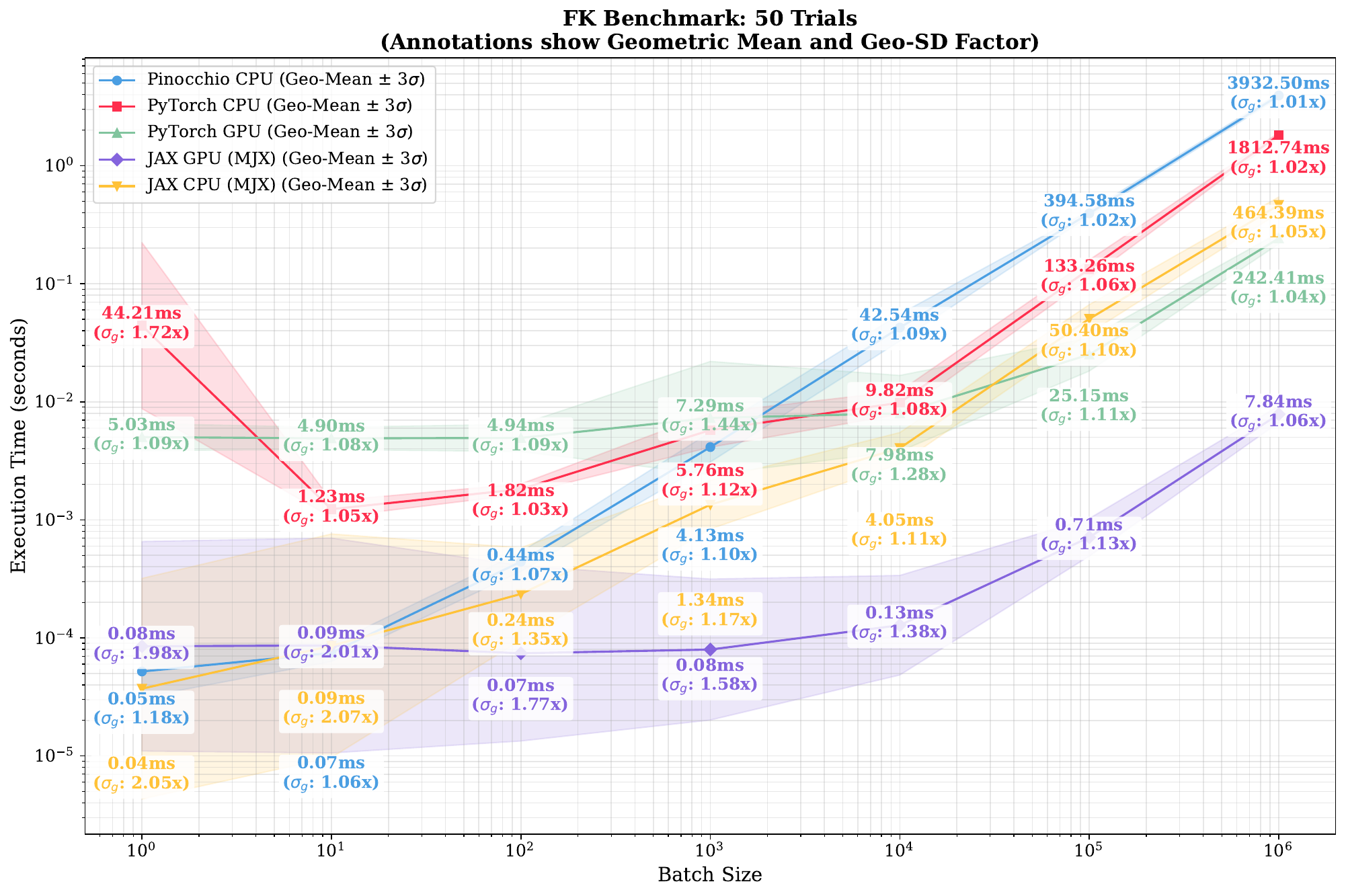}
    \caption{\textbf{Performance comparison of Python forward kinematics (FK) libraries.} We compared the performance of the following Python FK libraries: Pinocchio ~\cite{Carpentier2019_Pinocchio}, PyTorch Kinematics ~\cite{Zhong2024_PyTorchKinematics}, and JAX MJX ~\cite{jax2018github, Todorov2012_Mujoco, GoogleDeepmind_Mujoco}. Profiling execution times across different batch sizes. The superior parallel processing capabilities of JAX with MJX motivated its selection for our retargeter's core computations.}
    \label{fig:fk_benchmarks}
\end{figure}

Because of the heavy computational demands inherent to sampling-based approaches, it was critical to optimize our infrastructure to evaluate many samples in parallel across numerous update cycles. To achieve this, we profiled the latency and performance trade-offs of various Python Forward Kinematics (FK) libraries: Pinocchio ~\cite{Carpentier2019_Pinocchio}, PyTorch Kinematics ~\cite{Zhong2024_PyTorchKinematics}, and JAX MJX ~\cite{jax2018github, Todorov2012_Mujoco, GoogleDeepmind_Mujoco}. This benchmarking definitively motivated our decision to utilize JAX in conjunction with MJX for all FK computations, as it provided the necessary speed even at large batch sizes. All metrics were evaluated using a URDF of the mimic hand \cite{Nava2025_mimicone} on a workstation equipped with a 13th Gen Intel Core i9, 64 GB RAM, and an NVIDIA RTX 4090 (refer to \cref{fig:fk_benchmarks}). 

\begin{figure}[htbp]
    \centering
    \includegraphics[width=0.9\linewidth]{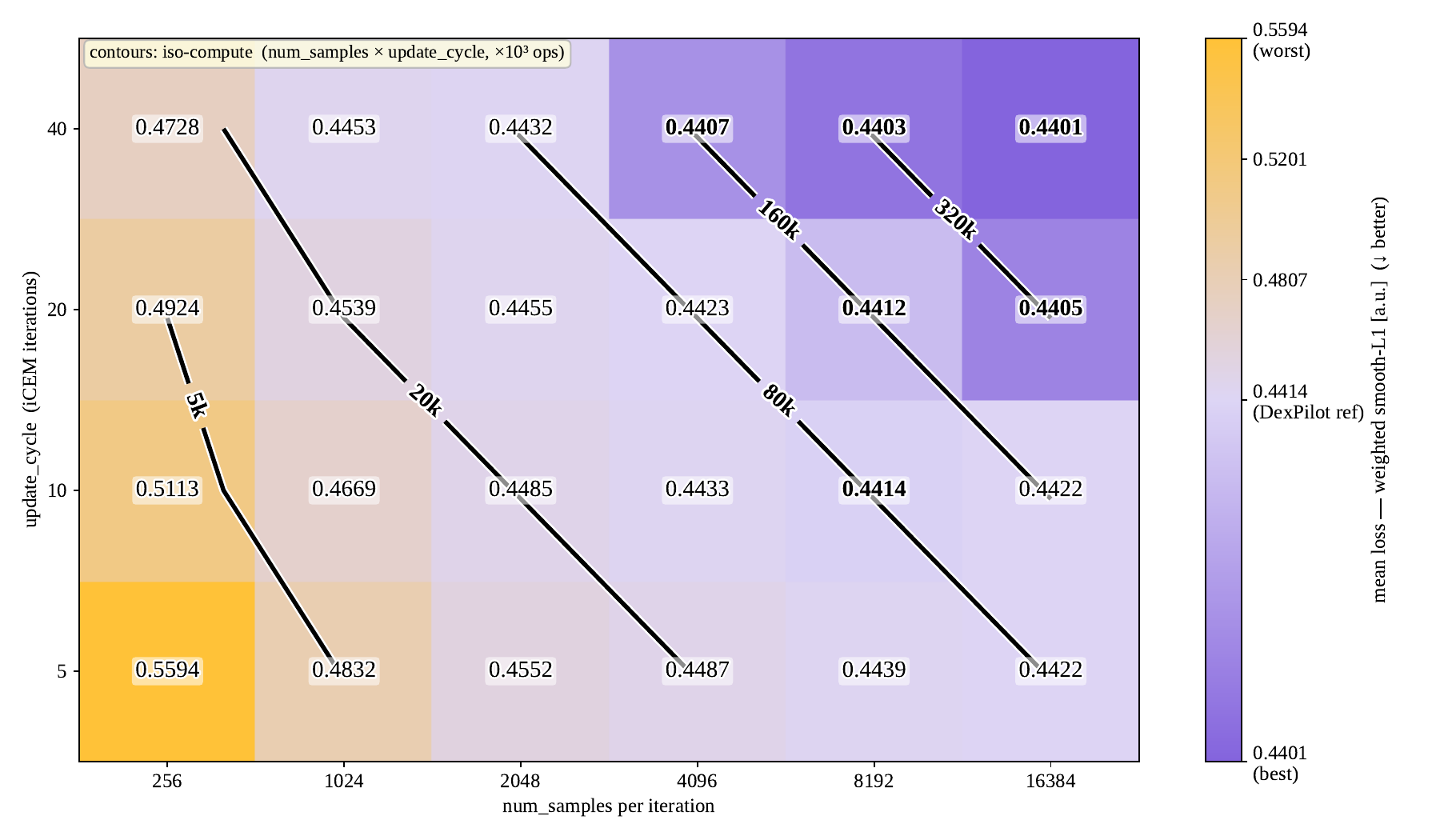}
    \caption{\textbf{Compute-performance trade-offs in SBR.} This ablation maps the mean tracking loss (weighted smooth-L1) across varying sample sizes $N$ and update cycles $L$. The diagonal contour lines denote equivalent compute budgets ($N \times L$). The green regions highlight hyperparameter configurations where SBR successfully surpasses the DexPilot reference loss ($0.4414$). Notably, the 160k operations contour reveals that a balanced allocation (e.g., $N=4096, L=40$ or $N=8192, L=20$) yields better optimization than heavily skewed sample generation ($N=16384, L=10$), emphasizing the need for both exploratory breadth and iterative depth.}
    \label{fig:offline_analysis_compute_performance_tradeoffs}
\end{figure}

With the underlying FK computation optimized, we then ablated the core SBR sampling hyperparameters (specifically the sample size $N$ and the number of update cycles $L$) to balance accuracy against latency constraints. As shown by the iso-compute contours $N \times L$ in \cref{fig:offline_analysis_compute_performance_tradeoffs}, successfully surpassing the DexPilot baseline tracking loss ($0.4414$) requires a balanced distribution of compute rather than blindly scaling a single parameter. Applying this optimal, balanced configuration over 2,000 retargeting calls yielded a highly efficient average latency of \textbf{3.664 ms}.

The complete configuration utilized for both the sampling and loss function parameters are provided in \cref{tab:sbr_hyperparams}. 

\begin{table}[htbp]
    \centering
    \caption{\textbf{Sampling-Based Retargeter (SBR) Hyperparameters.} The complete configuration used for the optimization engine and the task-specific loss formulation.}
    \label{tab:sbr_hyperparams}
    \begin{tabular}{lc}
        \toprule
        \textbf{Parameter} & \textbf{Value} \\
        \midrule
        \multicolumn{2}{c}{\textit{Sampling}} \\
        \midrule
        Number of samples ($N$) & $8192$ \\
        Elite samples ($M$) & $128$ \\
        Update cycles ($L$) & $20$ \\
        Initial covariance ($\sigma$) & $0.2$ \\
        Temperature ($\lambda$) & $0.1$ \\
        \midrule
        \multicolumn{2}{c}{\textit{Loss Formulation}} \\
        \midrule
        Scale factor ($\beta$) & $1.0$ \\
        Project distance [m] & $0.01$ \\
        Escape distance [m] & $0.015$ \\
        Primary vector weight ($w_{\text{set1}}$) & $2.0$ \\
        Secondary vector weight ($w_{\text{set2}}$) & $0.0$ \\
        Default vector weight ($w_{\text{default}}$) & $1.0$ \\
        Primary closing distance, $\eta_1$ [m] & $0.001$ \\
        Secondary separation distance, $\eta_2$ [m] & $0.0$ \\
        Joint centering weight ($\gamma$) & $0.05$ \\
        Velocity penalty weight ($w_v$) & $0.1$ \\
        \bottomrule
    \end{tabular}
\end{table}

%% file: thesis_smooth_operator/appendix/02_experiment_design.tex
\section{Experiment Design}
\label{sec:experiment_design}

\begin{figure}[htbp]
    \centering
    \includegraphics[width=0.35\textwidth]{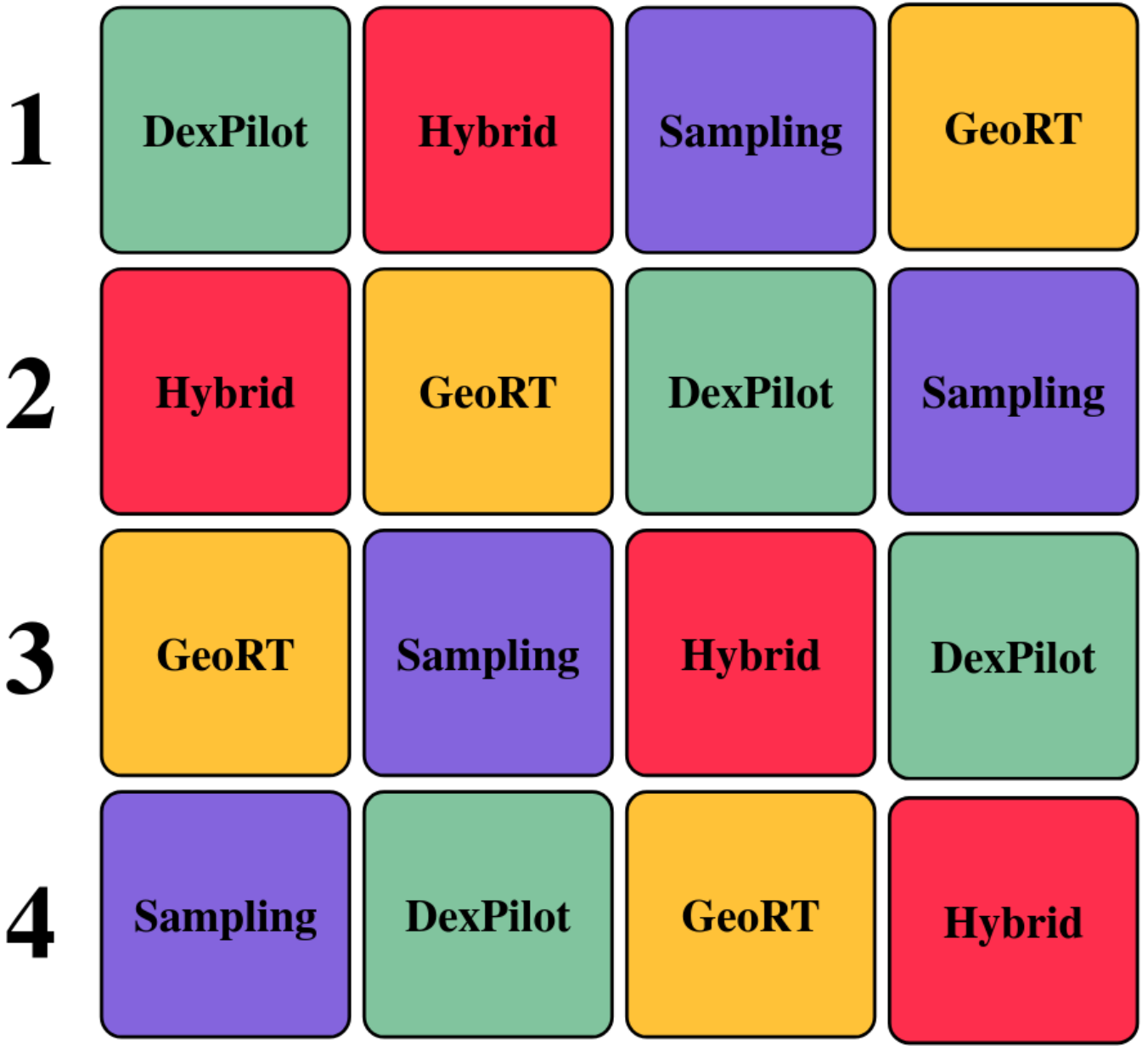} 
    \caption{\textbf{Balanced Latin Square matrix for counterbalancing sequence order.} To mitigate learning and fatigue carryover effects, operators were assigned to one of four sequence groups (rows). Each group evaluated the four retargeting algorithms in a strict chronological order (columns). This mathematical structure ensures that every retargeter appears in every ordinal position exactly once, and immediately follows every other retargeter exactly once.}
    \label{fig:balanced_latin_squares}
\end{figure}

\paragraph{Balanced Latin Squares}
\label{subsec:balanced_latin_squares}
To isolate the evaluation of each retargeting method from systemic order effects and progressive operator fatigue, our user study employs a Balanced Latin Square experimental design (\cref{fig:balanced_latin_squares}). The cohort of 18 operators is systematically distributed across counterbalanced evaluation sequences. Within an isolated task block, participants test four algorithms in a strict, pre-determined sequence: DexPilot~\cite{Handa2019_DexPilot}, GeoRT~\cite{Yin2025_GeoRT}, Hybrid~\cite{Nava2025_mimicone}, and SBR (Ours). For example, operators assigned to a specific group sequence execute their trials using GeoRT first, followed by SBR, Hybrid, and concluding with DexPilot. To maintain counterbalancing integrity across the entire study, operators evaluating multiple dexterous manipulation tasks are assigned to a distinct, non-overlapping Latin Square group sequence for each subsequent task. This structural control ensures that sequential carryover effects and operator adaptation trends are evenly distributed, allowing our statistical mixed-effects models to decouple underlying algorithmic performance from user familiarity.

\paragraph{Trial Preparation}
\label{subsec:trial_preparation}
Prior to testing, a standardized baseline calibration protocol is executed solely for the MANUS Metagloves Pro to normalize joint measurements across varying operator hand sizes. This calibration strictly adheres to the default skeletal profiling procedures outlined in the manufacturer's software interface \cite{ManusMetaglovesPro}. Crucially, to assess the true out-of-the-box performance and generalizability of each method on a standardized morphology, we do \textbf{not} perform any operator-specific fine-tuning or individual optimization of the Kabsch-Umeyama keypoint alignment transform or any internal retargeter hyperparameters. An identical, uniform global configuration and a fixed hardware control rate of 30Hz are maintained across all operators and trials.

Before recording data for any given retargeter block, operators are allocated a strict two-minute familiarization window. During this period, participants are permitted to move their fingers freely to gauge kinematic responsiveness but are explicitly forbidden from practicing the designated manipulation task or any variant of it. This protocol is specifically designed to capture immediate human-in-the-loop adaptation and learning curves. While prior works \cite{Naughton2024_ResPilot} allow operators to practice tasks prior to data recording, such a paradigm introduces a severe experimental confounding factor. Pre-trial practice obfuscates the evaluation by making it difficult to discern whether successful execution stems from an operator's manual compensation for systemic mapping errors or from the intrinsic fidelity, smoothness, and intuitiveness of the retargeting algorithm itself. This is exacerbated at small sample sizes. 

\paragraph{Task Success Rubric}
\label{subsec:task_success_rubric}
To provide a granular assessment of teleoperation proficiency, trial performance is quantified using a discrete scoring rubric: \textbf{1.0} point is awarded for perfect task completion, \textbf{0.5} points for marginal success, and \textbf{0.0} points for total task failure, which is defined as the object falling out of the hand's workspace bounds. Across all manipulation tasks, an operator is permitted to request a single manual object reset to the initial state at any point during an episode. Doing so automatically caps the maximum achievable score for that specific episode at \textbf{0.5} points. The precise physical conditions governing a marginal success score for each distinct task are structured as follows:

\begin{itemize}[leftmargin=*]
    \item \textit{Card Pickup (Task A)}: A score of \textbf{0.5} points is assigned if the card migrates out of the hand's immediate reachable workspace or is flicked across the top surface of the platform without moving over the edge, requiring a manual position reset to execute the final pinch gesture.
    \item \textit{Vertical Cube Rotation (Task B)}: A marginal score is triggered if an operator requests a manual reset due to an extreme misalignment of the tracking cube's faces. Alternatively, a score of \textbf{0.5} points is awarded without a reset if the operator successfully rotates the object but violates pure vertical rotation axis constraints, causing the cube to inadvertently flip onto an orthogonal face.
    \item \textit{Screwdriver Pivot (Task C)}: Unlike the previous environments, an intermediate milestone was not explicitly engineered into Task C. Instead, a score of \textbf{0.5} points was exclusively awarded when an operator successfully pivoted the tool close to the target downward vertical orientation but reached a terminal state or kinematic limit where no further fine adjustments could be executed to achieve perfect alignment.
\end{itemize}

\paragraph{Statistical Models}
\label{subsec:appendix_stat_models}
The real-world evaluation followed a within-subject, balanced Latin-square experimental design. Because participants performed tasks sequentially across all retargeting configurations, observations are clustered within individual operators, violating standard independent and identically distributed (i.i.d.) assumptions. To account for this dependency, we utilized the \texttt{statsmodels} library~\cite{Seabold2010_Statsmodels} to fit statistical models tailored to each metric.

\begin{itemize}[leftmargin=*]
    \item \textbf{Linear Mixed Models (LMM):} Completion Time and Overall Workload (NASA-TLX) were analyzed using LMMs~\cite{Gelman2006_DataAnalysisMEMs, Baayen2008_MixedEffectsModeling}. For the pooled overall model incorporating all tasks, the formulation is:
    \begin{equation}
      y_{ijk} = \mu + \alpha_{j} + \gamma_{k} + \beta\tilde{t}_{ijk} + u_{i} + \varepsilon_{ijk}
      \label{eq:lmm-pooled}
    \end{equation}
    where $\mu$ is the baseline intercept, $\alpha_{j}$ is the fixed effect of the retargeter, and $\gamma_{k}$ absorbs task-specific variance. To isolate within-subject operator learning curves and fatigue over time, $\beta\tilde{t}_{ijk}$ controls for the cumulative trial sequence index. Crucially, to handle between-subject variance, $u_{i} \sim \mathcal{N}(0,\sigma^{2}_{u})$ acts as a random intercept per operator to mathematically control for baseline dexterity differences across our cohort. $\varepsilon_{ijk}$ represents the residual error. Individual per-task models were fit identically, omitting the task covariate $\gamma_{k}$.

    \item \textbf{Logistic Generalised Estimating Equations (GEE):} Success Rate (trial score $\geq 0.5$) was modeled using a population-averaged logistic GEE rather than a subject-specific LMM~\cite{Gelman2006_DataAnalysisMEMs}. This ensures that the estimated performance probabilities are marginal (averaged over the entire operator population) rather than artificially conditioned on a specific individual's baseline skill level:
    \begin{equation}
      \operatorname{logit}(P(s_{ijk}=1)) = \mu + \alpha_{j} + \gamma_{k} + \beta\tilde{t}_{ijk}
      \label{eq:gee}
    \end{equation}
    where $s_{ijk}$ represents the dichotomized success outcome. 

    \item \textbf{Significance Testing:} Following model optimization, we calculated the Estimated Marginal Means (EMMs) for each retargeter. To verify baseline differences, each candidate retargeter was directly compared against the DexPilot baseline using a single-step Dunnett correction, which rigorously accounts for shared baseline dependency when comparing multiple treatments to a single control. For unconstrained, pairwise comparisons between all remaining retargeter combinations, significance values were adjusted using the Holm-Bonferroni procedure.
\end{itemize}

%% file: thesis_smooth_operator/appendix/03_operator_adaptation_effects.tex
\section{Operator Task Adaptation and Learning Effects}
\label{sec:operator_adaptation}

To accurately quantify the effect of online adoption experienced by human operators during real-world experiments, we aggregated performance metrics across all tasks and operators. Because teleoperation involves a non-trivial learning curve, we analyzed adaptation trends at both the global level (cumulative practice across the task) and the local block level (continuous practice using a single retargeter). 

At the global trial level, the general trend indicates that operators become faster as they gain more cumulative experience with the system. On average, the time required to complete a task decreases by $0.25$ seconds per trial ($p=0.022$). While the global success rate also sees a marginal increase of $0.007\%$, we do not find this global success rate trend to be statistically significant.

However, when examining adaptation at the local block level (where an operator is adjusting to one specific retargeter), we observe a much clearer and statistically significant learning effect. As operators progress through a specific block, their completion times decrease by $0.08$ seconds per trial ($p=0.021$), and their success rates increase by $0.004\%$ per trial ($p=0.041$). 

These statistically significant local trends confirm that, regardless of the assigned physical task, human operators actively learn and adapt to the nuances of different retargeters. 

Consequently, our findings demonstrate that adoption effects must be rigorously accounted for when designing human-in-the-loop robotic studies. Even with mitigation strategies like Balanced Latin Squares \cite{Bradely1958_BalancedLatinSquares}, maintaining a sufficiently robust sample size is crucial to ensure that these operator learning curves do not obscure the true performance of the underlying retargeting algorithms.

\begin{figure}[htbp]
    \centering
    
    \begin{subfigure}{0.48\textwidth}
        \includegraphics[width=\textwidth]{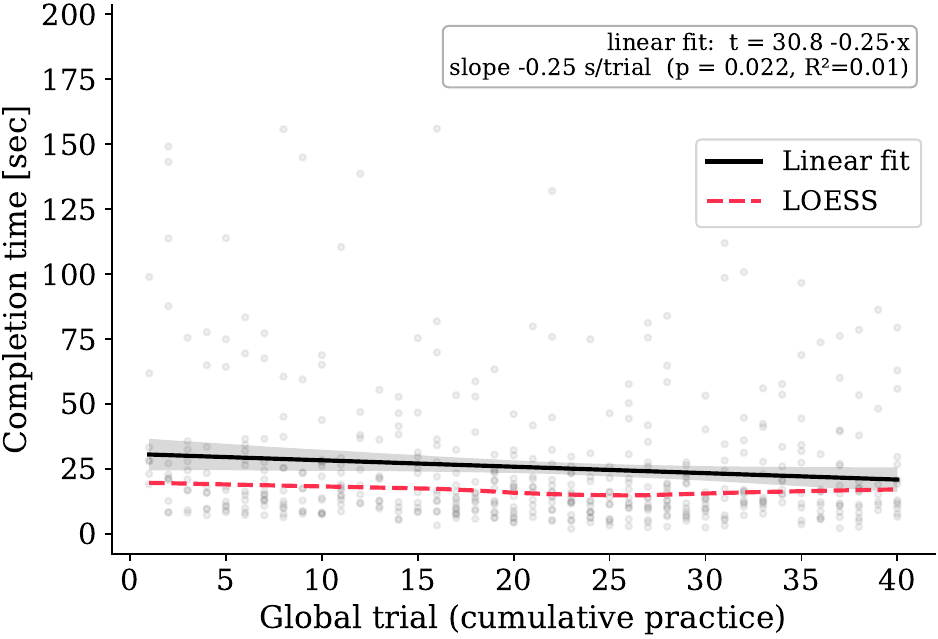}
        \caption{Completion Time (Global Trials)}
        \label{fig:learning_comp_global}
    \end{subfigure}
    \hfill
    \begin{subfigure}{0.48\textwidth}
        \includegraphics[width=\textwidth]{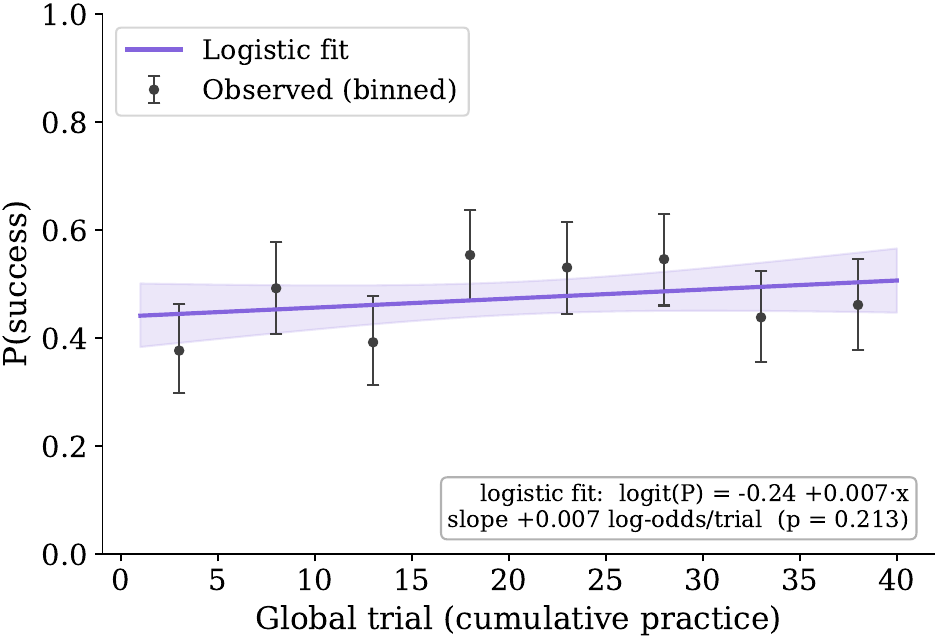}
        \caption{Success Rate (Global Trials)}
        \label{fig:learning_succ_global}
    \end{subfigure}
    
    \vspace{1em} 
    
    \begin{subfigure}{0.48\textwidth}
        \includegraphics[width=\textwidth]{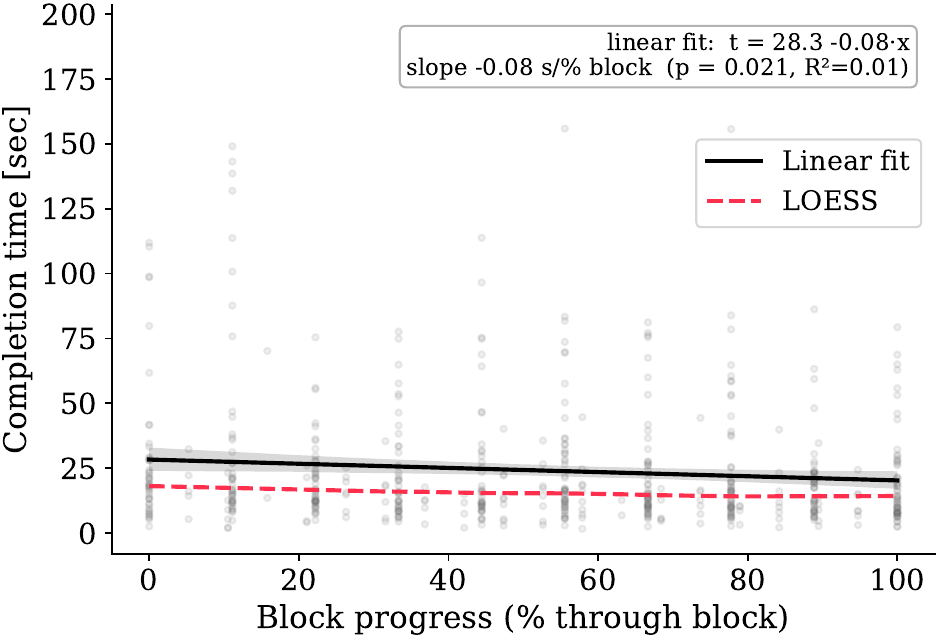}
        \caption{Completion Time (Local Trials)}
        \label{fig:learning_comp_block}
    \end{subfigure}
    \hfill
    \begin{subfigure}{0.48\textwidth}
        \includegraphics[width=\textwidth]{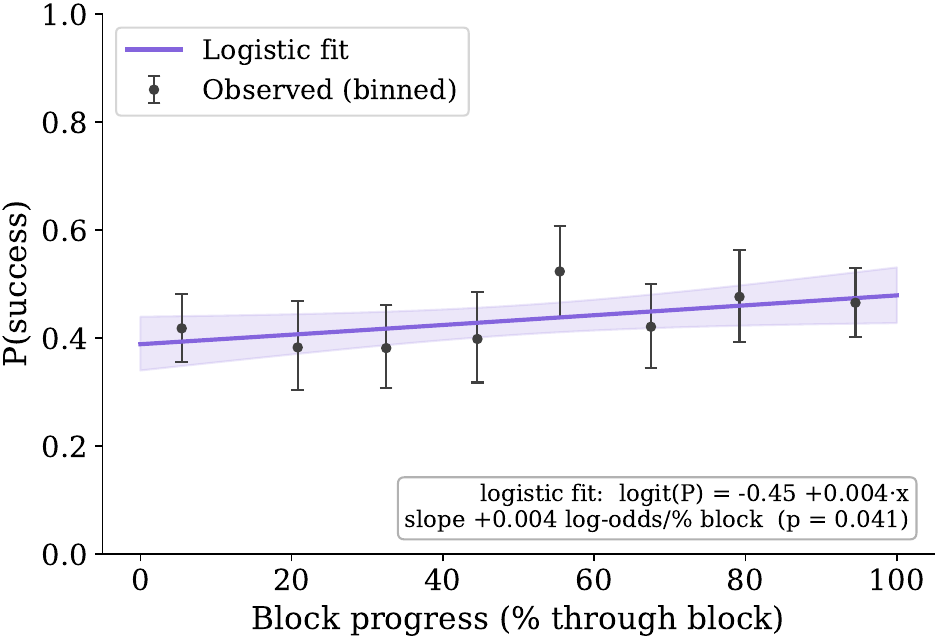}
        \caption{Success Rate (Local Trials)}
        \label{fig:learning_succ_block}
    \end{subfigure}
    
    \caption{\textbf{Overall operator task adaptation.} Aggregating all tasks and operators to quantify learning effects. At the global level, operators generally become significantly faster \textbf{(a)} as the study progresses. While success rates also trend upward globally \textbf{(b)}, this effect is not statistically significant. Conversely, at the local block level, significant adaptation effects emerge. This highlights that operators quickly adapt to the specifics of an individual retargeter over the course of a single block, as shown by statistically significant lower completion times \textbf{(c)} and increased success rates \textbf{(d)}.}
    \label{fig:overall_adaptation}
\end{figure}

%% file: thesis_smooth_operator/appendix/04_operator_variance.tex
\section{Operator Variance Analysis}
\label{sec:operator_variance}

\begin{figure}[htbp]
    \centering
    \includegraphics[width=1.0\textwidth]{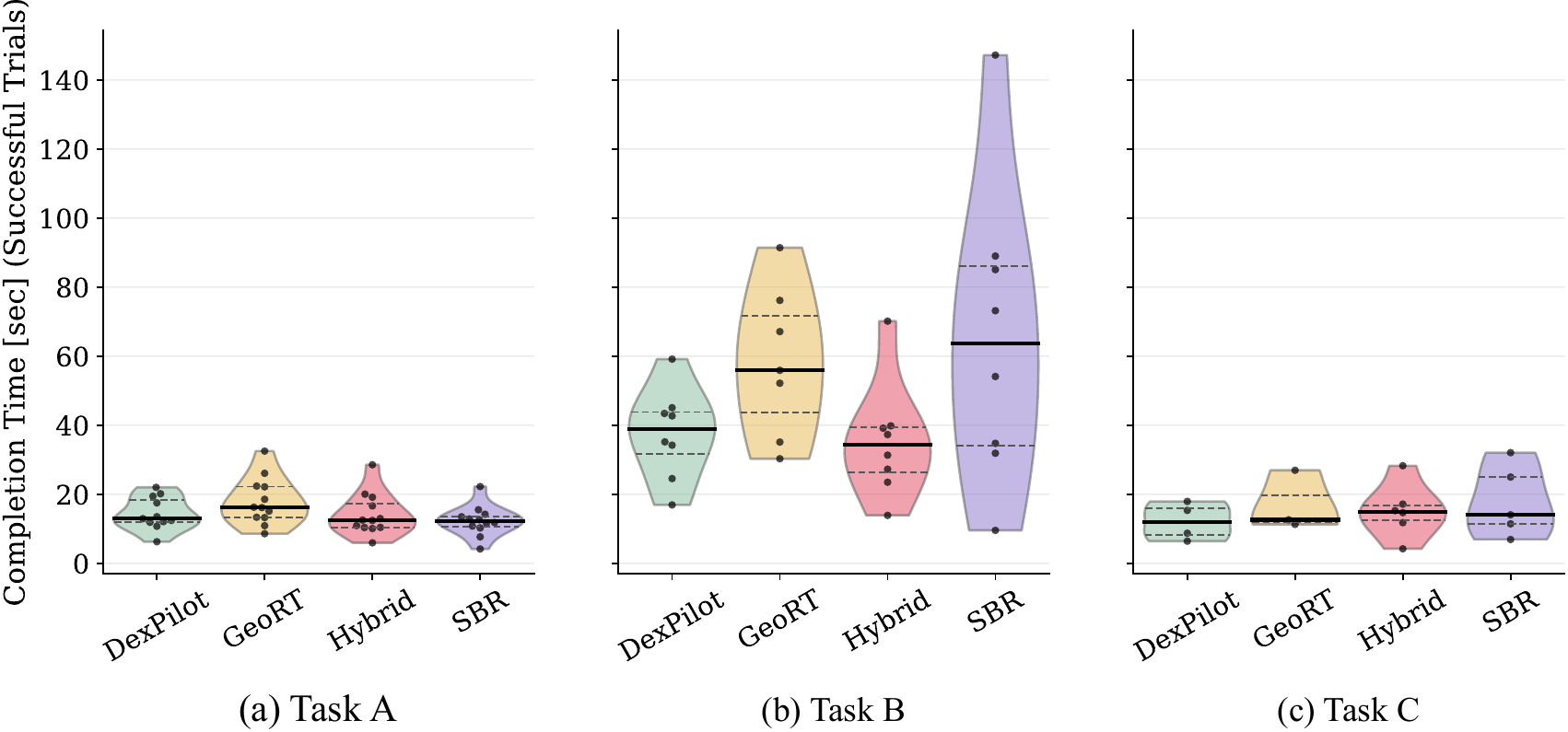} 
    \caption{\textbf{Operator variance in completion time across tasks.} Distribution of execution times across the four evaluated retargeting algorithms for the three tasks. Data points represent individual successful trials, with internal horizontal lines denoting the median values.}
    \label{fig:variance_time}
\end{figure}

\begin{figure}[htbp]
    \centering
    \includegraphics[width=1.0\textwidth]{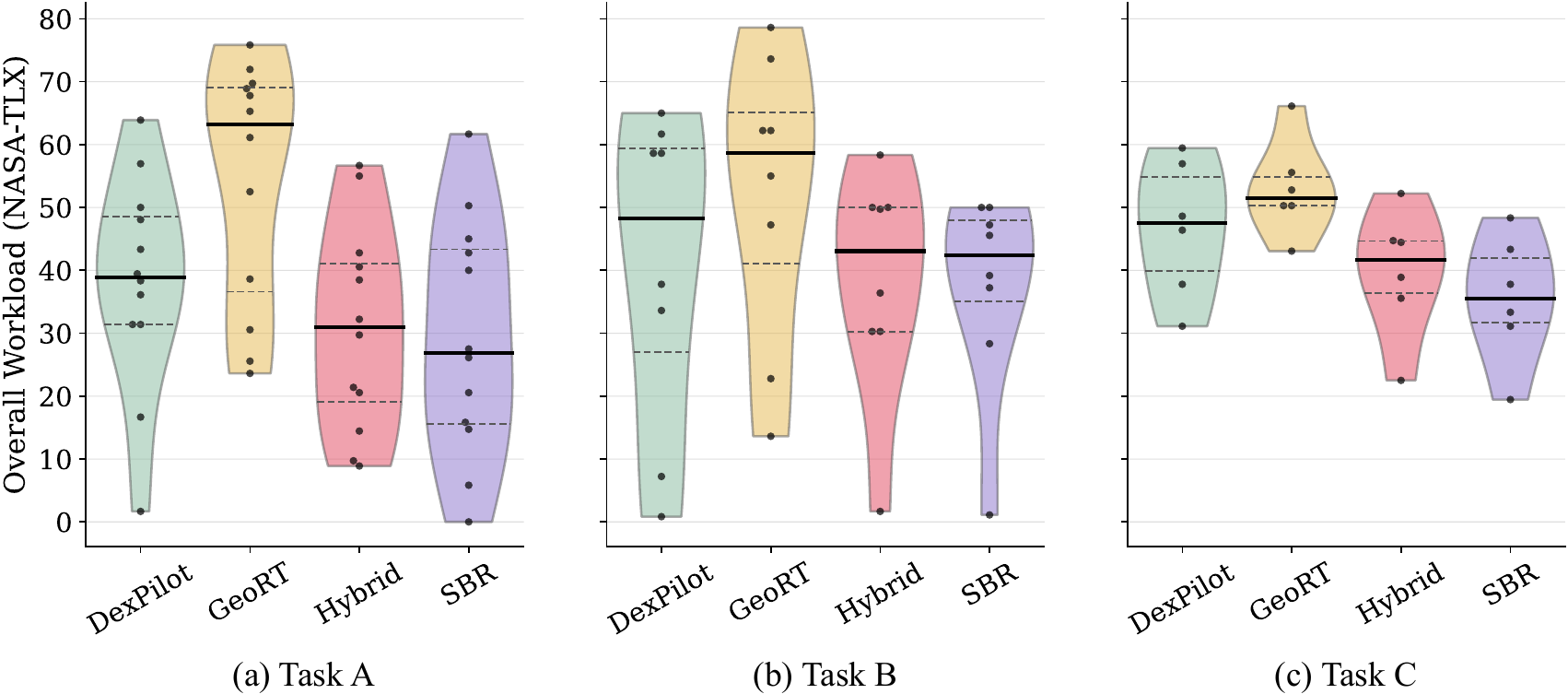} 
    \caption{\textbf{Operator variance in cognitive workload across tasks.} Total workload distributions derived via subjective NASA-TLX metrics across all tasks and retargeters. Despite the variance, we observe that SBR has the lowest median value (indicated by the solid black line) across the 3 different tasks, indicating that it has the lowest workload on the operator.}
    \label{fig:variance_workload}
\end{figure}

\begin{figure}[htbp]
    \centering
    \includegraphics[width=1.0\textwidth]{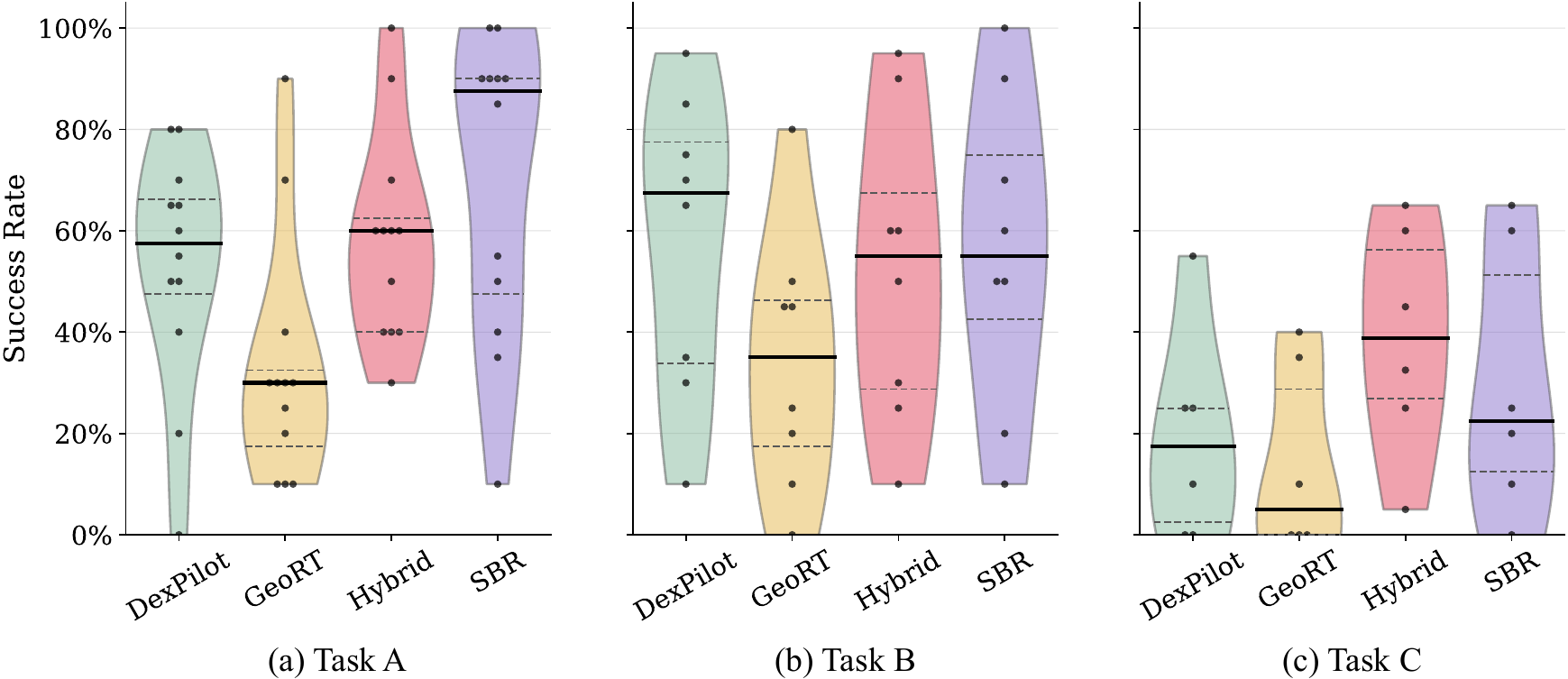} 
    \caption{\textbf{Operator success rate distributions across tasks.} Violin plots illustrating the spread of individual operator success rates across the three distinct manipulation tasks.}
    \label{fig:variance_success}
\end{figure}

To mitigate potential confounding factors such as learning curves and order effects, our real-world user study utilized a Balanced Latin Squares experimental design \cite{Bradely1958_BalancedLatinSquares}. Despite these structural controls, we observe substantial performance variance across the different human operators. This dispersion is primarily driven by two factors: the relatively compact sample size inherent to intensive physical human-in-the-loop testing, and the significant baseline skill gap separating individual operators. We analyze this operator-dependent variance across three key axes: completion times (\cref{fig:variance_time}), cognitive workload evaluated via the NASA-TLX protocol \cite{Hart1988_NASATLX} (\cref{fig:variance_workload}), and task success rates (\cref{fig:variance_success}).

As illustrated in \cref{fig:variance_time}, the completion times for Task A and Task C are noticeably shorter and more tightly bounded compared to Task B. Task B exhibits the widest overall variance across all four tested retargeters. This behavior stems directly from the mechanical nature of the task; to rotate the cube within the palm without dropping it, operators frequently choose to perform a series of slow, highly methodical micro-movements. This tactical divergence heavily inflates execution times for certain users while others complete the rotation rapidly, creating a wide spread in the data.

Beyond execution speed, user-dependent variations naturally extend into perceived cognitive demands and ultimately, task execution success (\cref{fig:variance_workload,fig:variance_success}). The distribution of overall workload and success rates reinforces the reality of operator diversity. While our participant cohort intentionally captures a broader, more heterogeneous mix of skill levels than what is typically reported in standard teleoperation literature, it exposes a critical methodological requirement for the robotics community. When evaluating interactive systems like robot retargeters, the extreme performance gaps between novice and expert operators mean that establishing a sufficiently large sample size is vital. Without it, individual user variance can easily mask the underlying algorithmic performance of the systems being tested. Providing these comprehensive variance profiles serves as an empirical baseline demonstrating why robust sample sizes are required to extract statistically definitive insights in teleoperation research.

%% file: thesis_smooth_operator/appendix/05_gradient_v_sampling.tex
\section{Gradient vs. Sampling-Based Optimization Ablation}
\label{sec:gradient_vs_sampling}

To better understand more theoretically the difference in performance between gradient-based and sampling-based approaches, we perform this ablation study. To perform this analysis, we utilize the DexPilot \cite{Handa2019_DexPilot} loss function, which is configured as follows:
\begin{equation}
\label{eq:dexpilot_loss_function}
    J_{DexPilot} = \sum_{k=1}^{K} w_{k}|| v_{R}^{k} - v_{H}^{k}||^2 + w_{reg}|| q_{R} - q_{R,neutral}||^2
\end{equation}

where $K$ is the number of keyvectors, $q_{R}$ is the joint angles of the robot hand and $q_{H}$ is the pose of the human hand. $v_{H}^{k}$ and $v_{R}^{k}$ represent the human and robot keyvectors, respectively. $q_{R, neutral}$ is the neutral hand pose (defined as zero angles). $w_k$ is the weight for the $k$-th keyvector term, while $w_{reg}$ represents the weights of the pose regularization and velocity regularization terms. Importantly, to properly ablate and isolate performance drivers, we only change the underlying optimization approach while keeping all other components identical. 

Additionally, to give a fair comparison with the numerous samples that are being tested by the sampling-based approach, we implement a batch optimization framework where we run $N$ optimization steps in parallel and apply a softmax on their outputs. Specifically, we use $N=32768$ samples for this batch optimization, directly matching the count utilized in the sampling-based method. For our baselines, RMSProp \cite{Tieleman2012_RMSProp} was chosen since it is a non-momentum based method, and SLSQP \cite{Johnson_SLSQP} was selected because it is commonly used by conventional retargeters in the literature \cite{Handa2019_DexPilot, Xin2025_AnalyzingKeyObjectives}. Because of the extreme computational demands of running SLSQP in parallel, we limited this part of our analysis to 500 frames of recorded hand motion, which is equivalent to approximately 5 seconds of execution.

\begin{figure}[htbp]
    \centering
    \begin{subfigure}{0.495\textwidth}
        \includegraphics[width=\textwidth]{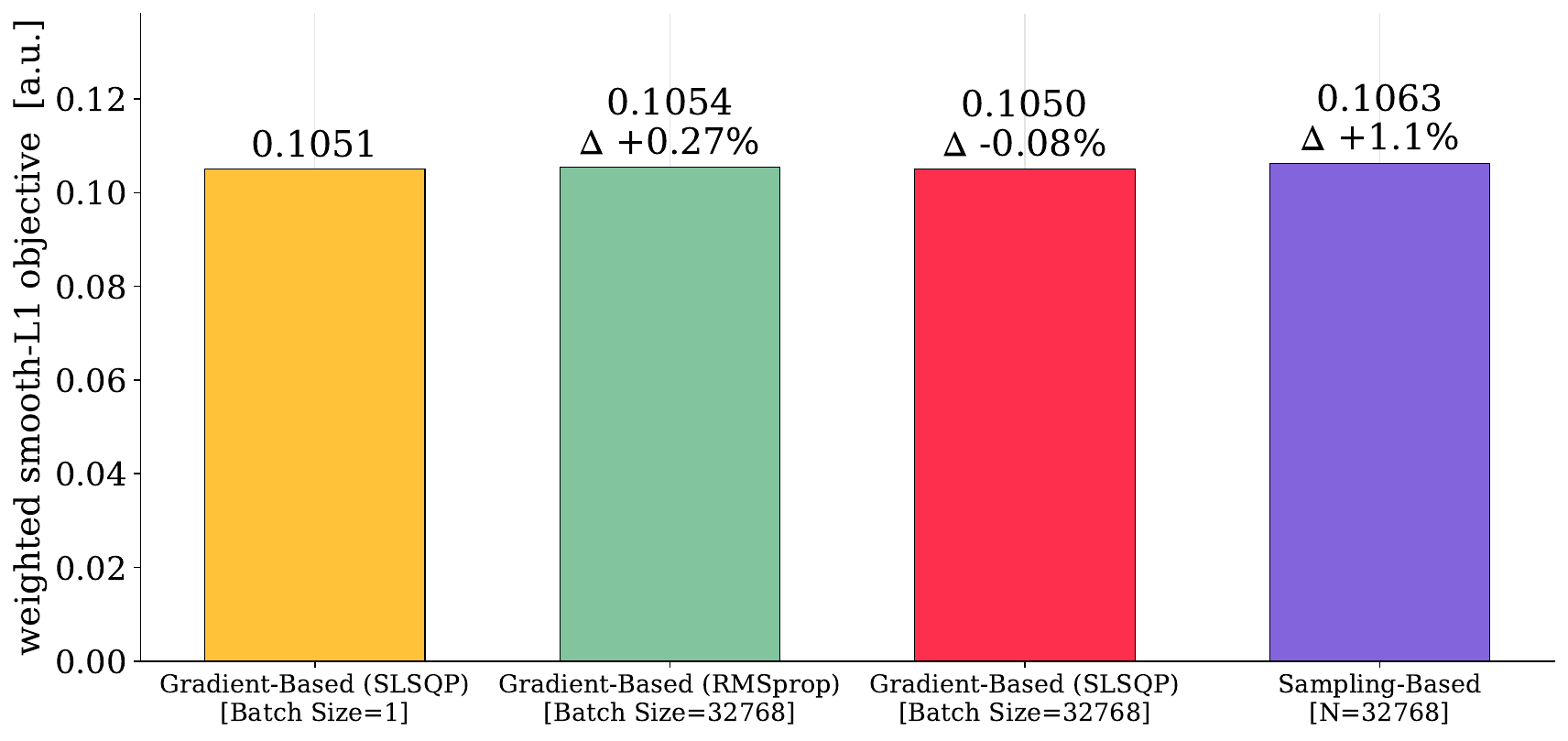}
        \caption{Mean Loss}
        \label{fig:grad_vs_sampling_loss}
    \end{subfigure}
    \hfill
    \begin{subfigure}{0.495\textwidth}
        \includegraphics[width=\textwidth]{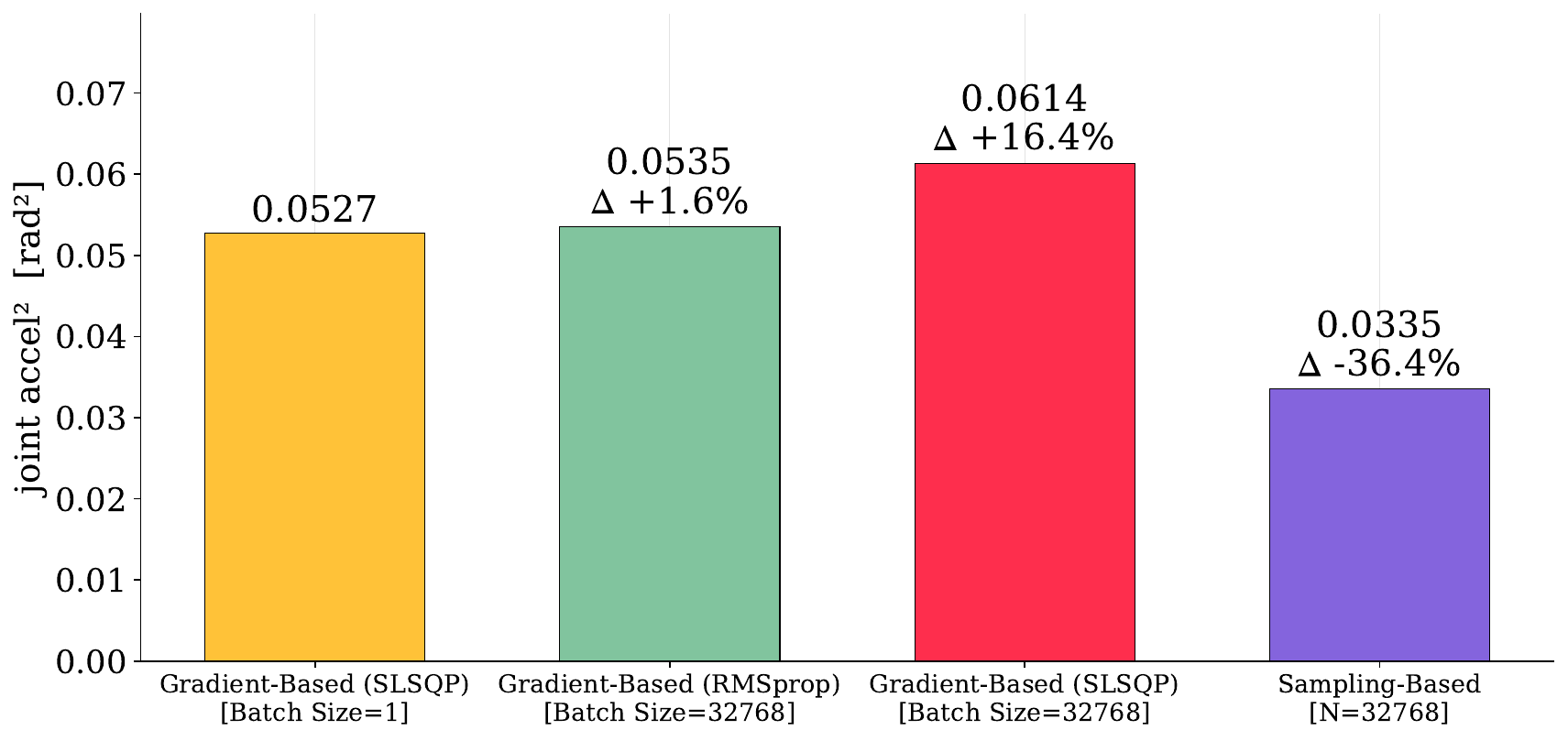}
        \caption{Joint Acceleration (Jitter)}
        \label{fig:grad_vs_sampling_jitter}
    \end{subfigure}
    
    \caption{\textbf{Gradient versus sampling-based optimization comparison.} Utilizing the DexPilot \cite{Handa2019_DexPilot} loss function and swapping out the optimization methods, we compare the baseline tracking performance. To further help with the comparison, we perform batch optimization where we run the optimized inputs in parallel and apply a softmax to their outputs using $N=32768$ samples, matching the sampling-based approach. RMSProp \cite{Tieleman2012_RMSProp} was used as a non-momentum baseline, and SLSQP \cite{Johnson_SLSQP} was selected as it is used among existing retargeters \cite{Handa2019_DexPilot, Xin2025_AnalyzingKeyObjectives}. Due to the heavy computational demands of parallel SLSQP execution, we analyze 500 frames of recorded hand motion ($\approx 5$~seconds). As shown in \textbf{(a)}, all methods achieve approximately identical mean loss values. However, the sampling-based approach achieves a $36.4\%$ reduction in tracking jitter \textbf{(b)} compared to the gradient-based SLSQP baseline.}
    \label{fig:grad_vs_sampling_comparison}
\end{figure}

\paragraph{Base Optimization Performance}
\label{subsec:base_optimization_performance}
Looking at the primary optimization results, we can see that while the final tracking losses are remarkably similar to the gradient-based baseline, the overall joint jittering is significantly reduced by $36.4\%$ when using the sampling-based approach. This performance breakdown is illustrated in \cref{fig:grad_vs_sampling_comparison}.

\paragraph{Hessian Eigen-Direction Analysis}
\label{subsec:hessian_eigendirection_analysis}
To better understand why the sampling-based method outperforms the gradient baselines in consistency, we look at the eigen-directions of the Hessian matrix. The low eigenvector ranks correspond to the flat regions of the loss landscape, indicated by the lower weight of their respective eigenvectors. Intuitively, this behavior emerges because in the null space, the loss landscape is completely flat. 

Gradient-based methods converge to and subsequently jump around these different valid solutions within the flat region. This issue is exacerbated even further if there is tracking noise on the input side from the hand tracker. However, sampling-based approaches are uniquely able to navigate this landscape safely since the optimization paradigm inherently acts as a low-pass filter.

\begin{figure}
    \centering
    \includegraphics[trim=0cm 0cm 0cm 1.0cm, clip, width=0.80\textwidth]{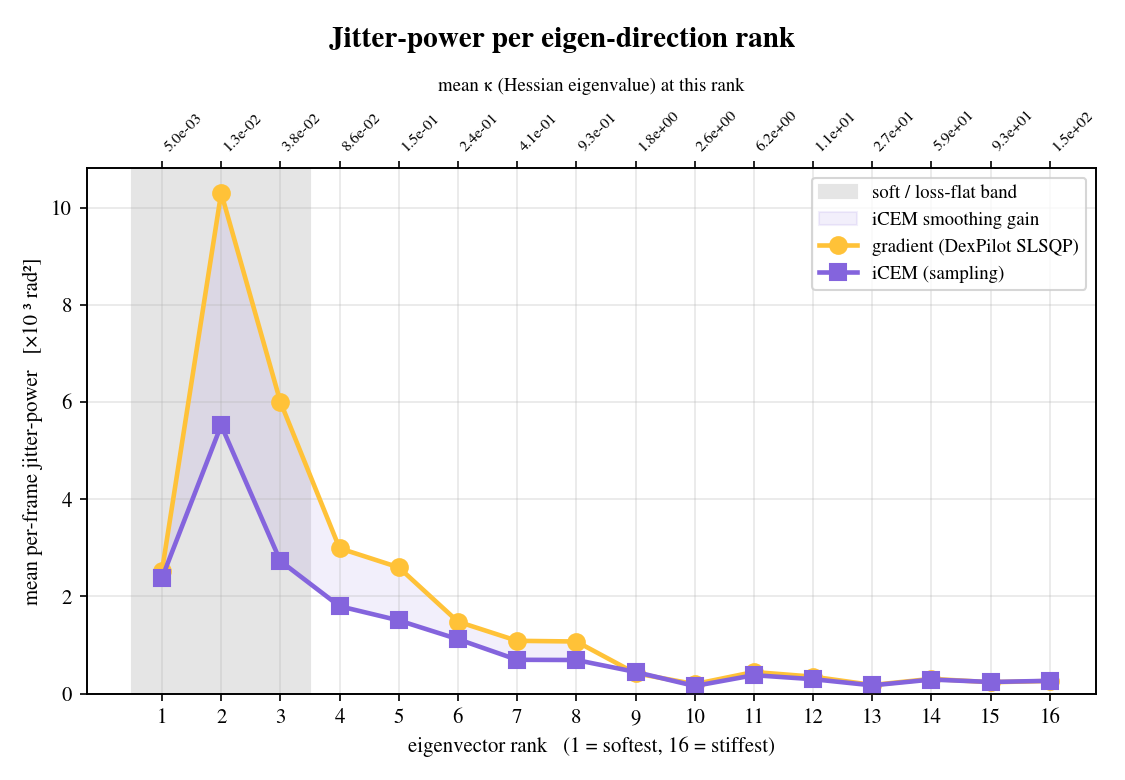} 
    \caption{\textbf{Hessian eigen-direction decomposition and comparison.} The eigen-modes of the Hessian enable us to better understand the true source of tracking jitter, given that jitter represents the second derivative with respect to position. For the stiffer eigenvectors (characterized by higher eigenvector and eigenvalue ranks), we observe that the mean per-frame jitter is very close across all methods. However, as we move towards the lower-ranked eigen-directions, we begin to notice a stark divergence. This trend indicates that in the flat regions of the loss function, the gradient-based method jumps around and fails to converge to a consistent value.}
    \label{fig:hessian_eigen}
\end{figure}

As visualized in \cref{fig:hessian_eigen}, for the stiffer eigenvectors (higher eigenvector and eigenvalue ranks), the mean per-frame jitter is closely matched. However, as we move down towards the lower-ranked eigen-directions, a significant divergence appears. This confirms that within the loss-flat bands, gradient-based methods suffer from erratic behavior.

\paragraph{Velocity Regularization Trade-offs}
\label{subsec:ablation_velocity_regularization}
To complete the optimization trade-off analysis, we also investigated the effects of explicitly introducing a velocity regularization term into the objective function. To do this, we optimize the same base loss function in \cref{eq:dexpilot_loss_function} but add a velocity regularization component $w_{v}|| q_{R}^{} - q_{R,prev} ||^2$, where $q_{R}$ is the current pose of the robot, $q_{R,prev}$ is the pose of the robot in the previous timestep, and $w_{v}$ is the weight of the velocity regularization term. We set $w_v = 0.05$.

\begin{figure}[htbp]
    \centering
    \begin{subfigure}{0.495\textwidth}
        \includegraphics[width=\textwidth]{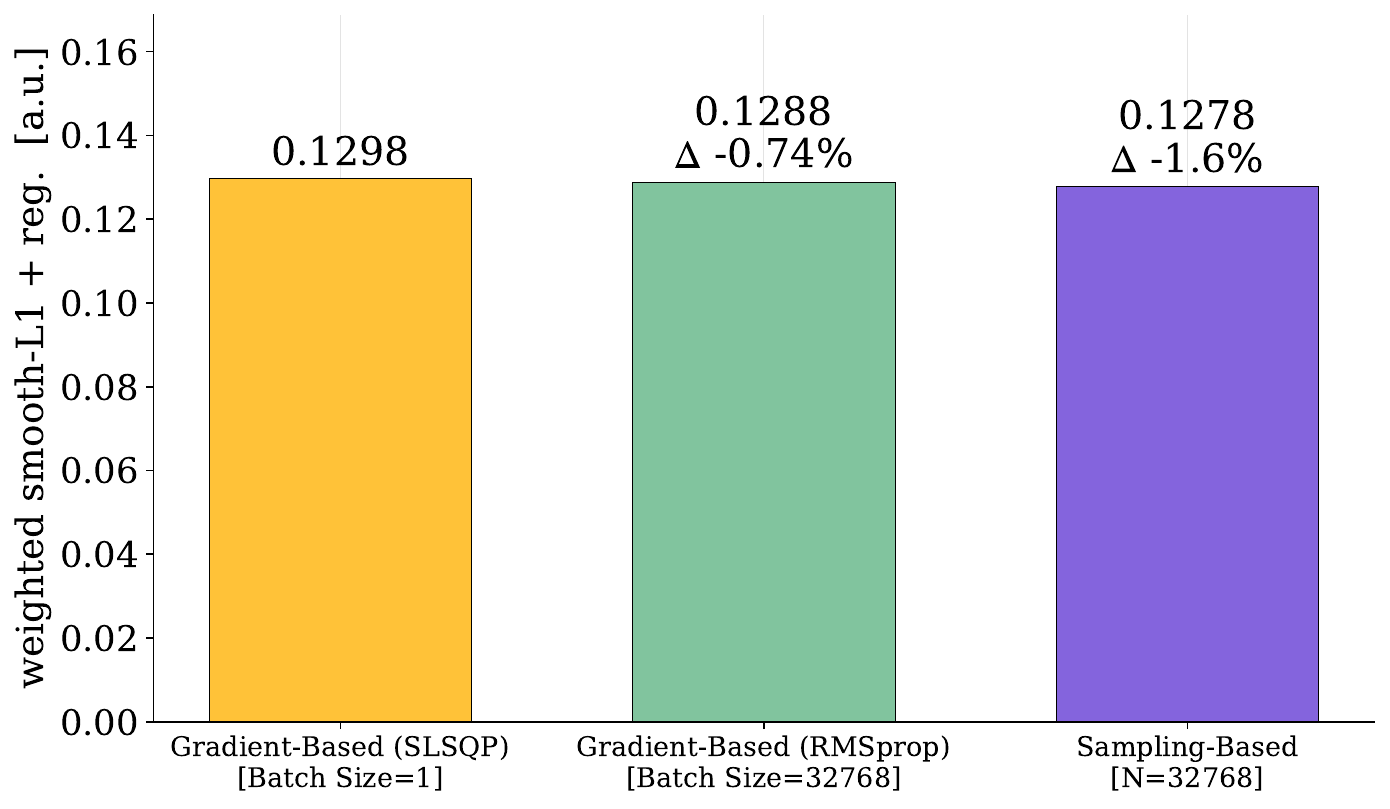}
        \caption{Mean Loss}
        \label{fig:velocity_reg_loss}
    \end{subfigure}
    \hfill
    \begin{subfigure}{0.495\textwidth}
        \includegraphics[width=\textwidth]{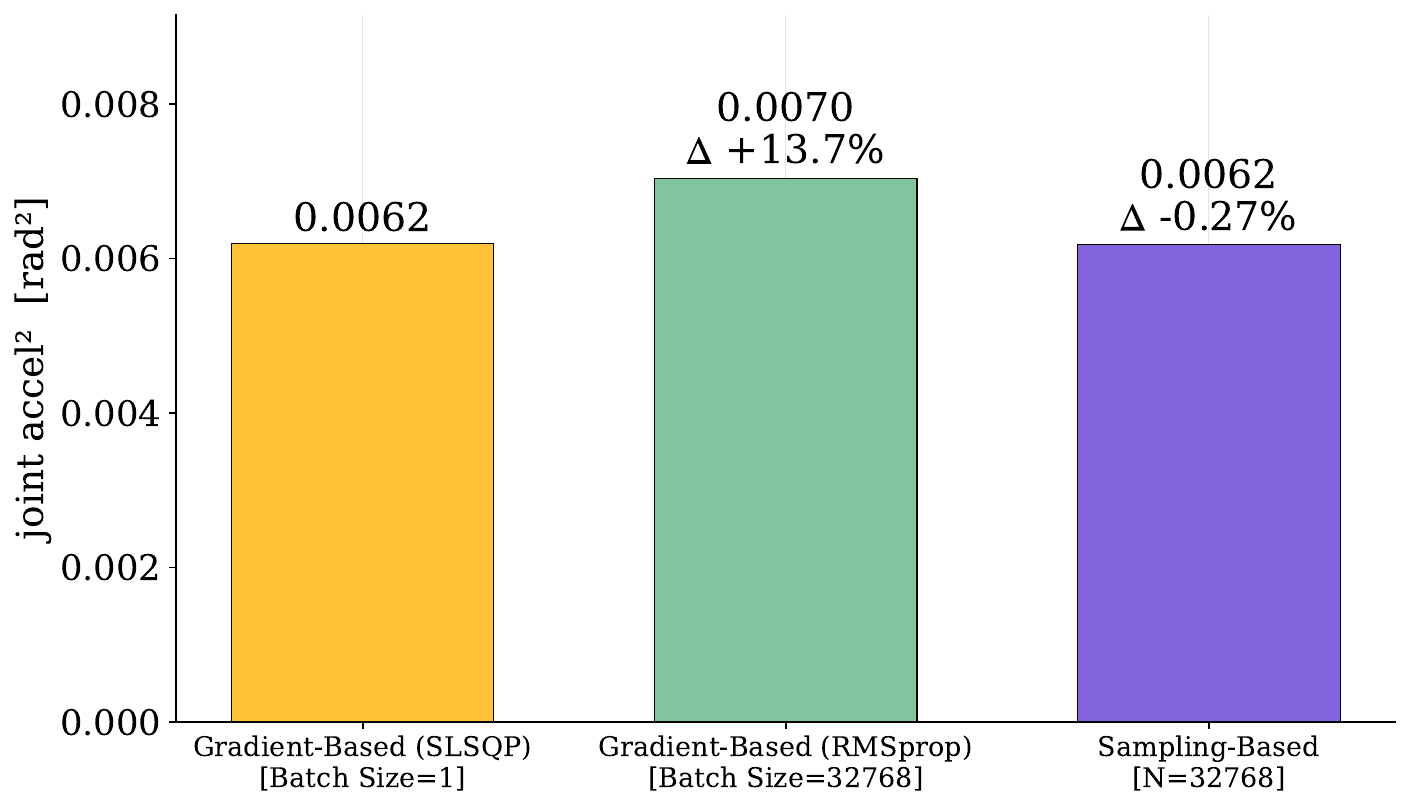}
        \caption{Joint Acceleration (Jitter)}
        \label{fig:velocity_reg_jitter}
    \end{subfigure}
    
    \caption{\textbf{Gradient- versus sampling-based optimization comparison with a velocity regularization term.} Using the same core DexPilot loss function, we introduce an explicit velocity regularization term with a weight of $w_v=0.05$. Although joint jittering is successfully reduced across the board by a factor of 7.18 times \textbf{(b)}, the mean tracking loss increases by 0.0232 on average \textbf{(a)}. Under this formulation, the gradient-based method achieves comparable jitter to sampling-based methods, but at the explicit trade-off of incurring slightly higher tracking loss costs.}
    \label{fig:velocity_reg}
\end{figure}

From the results shown in \cref{fig:velocity_reg}, adding the velocity regularization term successfully reduces tracking jitter across the board by approximately 7.18 times. However, this improvement comes with a notable trade-off: it increases the mean tracking loss by 0.0232 on average. While the gradient-based method can be forced to achieve the same low jitter profile as sampling-based methods using this penalty, it does so at the cost of incurring significantly higher overall loss metrics. 

Ultimately, this comprehensive ablation highlights that sampling-based approaches can natively achieve excellent, low-jitter tracking performance comparable to gradient-based methods without requiring heavy regularization penalties that degrade tracking accuracy.

%% file: thesis_smooth_operator/appendix/06_sim_to_real.tex
\section{Sim-to-Real Metric Correlation Analysis}
\label{sec:sim_to_real_metrics}

Because real-world robotics experiments are resource-intensive and expensive to conduct, establishing reliable simulation-based metrics that accurately predict physical performance is crucial. To address this, we evaluate a suite of proxy metrics drawn from the existing retargeting literature \cite{Handa2019_DexPilot, Yin2025_GeoRT, Naughton2024_ResPilot}. These metrics specifically capture key performance drivers along three primary axes: \textbf{Accuracy}, \textbf{Consistency}, and \textbf{Efficiency}. They are defined as below:

\begin{itemize}
    \item \textbf{Accuracy:} Keyvectors capture the intended spatial relationship between embodiments \cite{Handa2019_DexPilot, Qin2024_AnyTeleop}. We assess the directional alignment of corresponding human ($\mathbf{v}^k_H$) and robot ($\mathbf{v}^k_R$) keyvectors via \underline{Cosine Similarity} ($\frac{\mathbf{v}^k_H \cdot \mathbf{v}^k_R}{||\mathbf{v}^k_H|| ||\mathbf{v}^k_R||}$). This metric ranges from $[-1, 1]$, where $1$ indicates perfect parallel alignment and $-1$ indicates diametrically opposite directions. Spatial magnitude preservation is measured via the \underline{Scale Ratio} ($||\mathbf{v}^k_{R}|| / ||\mathbf{v}^k_{H}||$), where an optimal value approaching $1.0$ indicates minimal spatial distortion.
    
    \item \textbf{Consistency:} This quantifies kinematic stability, critical for operator intuitiveness \cite{Yin2025_GeoRT}. \underline{Motion Preservation} assesses the directional tracking of normalized human ($\Delta \bar{\mathbf{p}}_{H}^t$) and robot ($\Delta \mathbf{p}_{R}^t$) fingertip velocity displacements across a time step $t$: $\frac{\Delta \bar{\mathbf{p}}_{H}^t \cdot \Delta \mathbf{p}_{R}^t}{\|\Delta \bar{\mathbf{p}}_{H}^t\| \|\Delta \mathbf{p}_{R}^t\|}$. This metric ranges from $[-1, 1]$, where $1$ signifies identical movement direction and $-1$ signifies strictly opposing motion. \underline{Flatness} penalizes high-frequency control jitter by computing a discrete approximation of squared acceleration for the robot landmark trajectory $\mathbf{p}_{R}$: $\|\mathbf{p}_{R}^{t+1} + \mathbf{p}_{R}^{t-1} - 2\mathbf{p}_{R}^{t}\|^{2}$. Optimal flatness approaches $0$.
    
    \item \textbf{Efficiency:} We evaluate the retargeter's ability to leverage mechanical range while maintaining real-time frequencies. \underline{Workspace Utilization} measures the volumetric overlap between the reachable workspace \cite{Yin2025_GeoRT, Naughton2024_ResPilot}. This is approximated by taking the retargeted fingertip spheres (radius $r$, centered at human input $x_{H}^{i}$ mapped via forward kinematics $\operatorname{FK}$) and the robot's reachable workspace ($KP_{R}^{i}$): $\frac{\operatorname{Vol}(\cup_{i}\mathcal{B}(\operatorname{FK}\circ f(x_{H}^{i}),r) \cap KP_{R}^{i})}{\operatorname{Vol}(KP_{R}^{i})}$. This yields a coverage percentage in $[0\%, 100\%]$, where higher is better.
\end{itemize}

\begin{figure}
    \centering
    \includegraphics[trim=0cm 0cm 0cm 0.5cm, clip, width=0.95\textwidth]{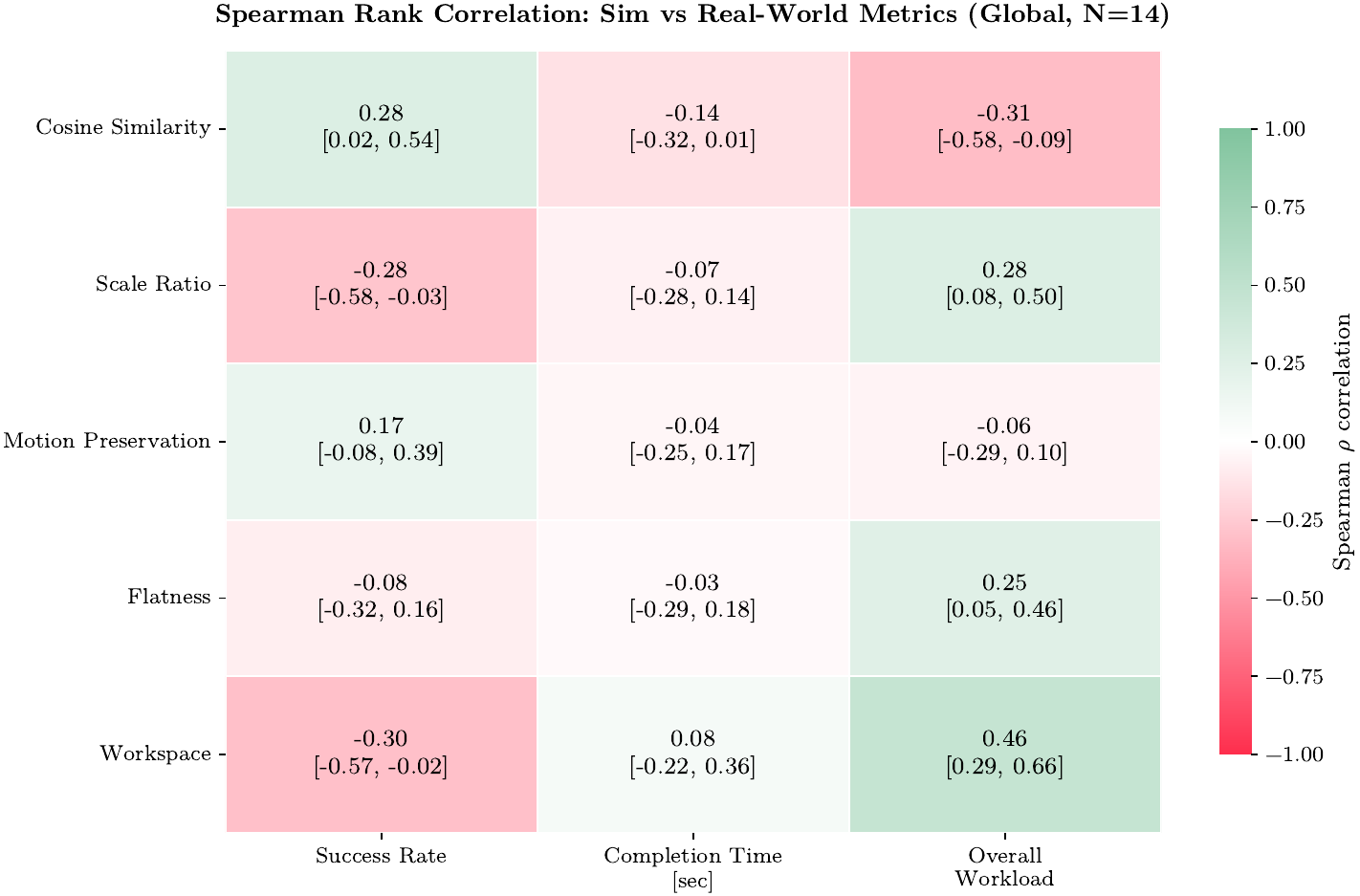}
    \caption{\textbf{Spearman rank correlation between simulation metrics and real-world performance.} Human-in-the-loop study results ($N=14$ operators) comparing simulation metrics (rows) against real-world metrics (columns). In accordance with standard behavioral science frameworks \cite{Cohen1988_EffectSize}, correlation magnitudes are interpreted relatively: $r=\pm0.1$ denotes a small effect, $r=\pm0.3$ a medium effect, and $r=\pm0.5$ a large effect. Confidence intervals (CIs) are reported in brackets. CIs that do not cross zero indicate statistical significance.}
    \label{fig:sim_to_real_correlation}
\end{figure}

To evaluate their predictive power, we compute the Spearman rank correlation coefficient ($\rho$) between these simulation-derived heuristics and observed real-world performance markers. The resulting correlation matrix is illustrated in \cref{fig:sim_to_real_correlation}.

Due to the inherent variability of human-in-the-loop experiments, these correlation magnitudes must be interpreted using relative benchmarks established in behavioral sciences \cite{Cohen1988_EffectSize}. Following Cohen's conventions \cite{Cohen1988_EffectSize}, a correlation of $r = \pm0.1$ indicates a small effect size, $r = \pm0.3$ represents a medium effect, and $r = \pm0.5$ reflects a large effect. Statistically significant relationships are identified by confidence intervals (CIs) that do not span or cross over zero.

\paragraph{Analysis of Heuristics Correlations}
\label{subsec:analysis_of_heuristics_correlations}
Analyzing the data, we make the following observations:
\begin{itemize}
    \item \textbf{Cosine Similarity:} Our data reveals that Cosine Similarity exhibits a statistically significant effect on both real-world success rates and overall cognitive workload. Specifically, it demonstrates a positive, near-moderate correlation with success rates ($r = 0.28$), demonstrating that closer directional alignment of keyvectors yields a higher probability of successful task execution. Conversely, a distinct, moderate negative relationship emerges with overall workload ($r = -0.31$), indicating that cognitive load significantly drops as the cosine similarity metric approaches $1.0$. Taken together, these insights strongly highlight that a retargeter's ability to preserve the human operator's original directional alignment is a vital determinant of teleoperation performance.

    \item \textbf{Scale Ratio:} The Scale Ratio metric exhibits a near-moderate inverse correlation ($r = -0.28$) with real-world success rates and a near-moderate direct correlation ($r=0.28$) with overall workload. This means that a lower scale ratio is associated with a higher likelihood of task success and reduced overall workload. This structural trend hints that a retargeting algorithm benefits more from overcompensating for human-to-robot scale differences ($< 1.0$) than under-compensating for them ($> 1.0$). However, this boundary condition should not be over-extrapolated; driving the scale ratio completely down toward $0.0$ is not a functional or ideal solution.

    \item \textbf{Flatness and Motion Preservation:} Motion Preservation demonstrates no statistically significant relationship with real-world success rates, completion times, or cognitive workload within this framework. On the other hand, Flatness presents no significant correlations \textit{except} for a modest positive association with overall workload ($r = 0.25$). This selective interaction is logical: when the tracking mapping is exceptionally smooth (approaching $0.0$), the immediate physical and cognitive workload imposed on the operator decreases.

    \item \textbf{Workspace:} Interestingly, Workspace displays a medium-sized inverse relationship with task success ($r = -0.30$) and a near large-sized effect with overall workload ($r=0.46$). This unexpected reversal is primarily driven by an anomaly between the two extreme baselines in our study: GeoRT \cite{Yin2025_GeoRT} and SBR. While GeoRT and SBR achieved the highest and lowest simulation workspace metrics respectively, their real-world performance flipped, with SBR dominating the top rankings and GeoRT performing poorly. Consequently, further targeted validation studies must be performed to better isolate and quantify this metric's true utility.
\end{itemize}

Notably, none of the evaluated simulation metrics show a statistically significant correlation with real-world completion times. Qualitative observations during our user studies suggest this flat response is heavily influenced by divergent user teleoperation styles: certain operators choose a slow, highly methodical strategy, whereas others prioritize speed and race to finish the task.

\paragraph{Discussion and Limitations}
\label{subsec:discussion_and_limitations}
While these findings offer essential directions for the domain, several constraints remain. First, these experimental metrics were swept across a limited population of four distinct retargeting algorithms. Evaluating a wider collection of methods is necessary to conclusively solidify the mathematical bridge between simulation proxies and reality.

Next is the sample size of operators. Due to participant scheduling constraints, only a subset of 14 operators completed the simulation analysis. These same logistical limitations account for the varying participant counts across the individual real-world experimental tasks. Although our cohort of $N=14$ physical operators is robust and sizable relative to the baseline standard in current retargeting literature, scaling up the number of human participants in future work will further validate these behavioral effects.

Ultimately, while these metrics are not a definitive, all-encompassing solution for sim-to-real evaluation, we present these correlation baselines to the community to serve as a foundational benchmark for developing more predictive, high-fidelity robotic retargeting metrics in the future.